\documentclass[runningheads]{llncs}
\usepackage{graphicx}
\usepackage{amsmath,amssymb}
\usepackage{color}

\usepackage{subfig}
\usepackage{tabularx}
\usepackage{booktabs}
\usepackage{multirow}
\usepackage{makecell}
\usepackage{xcolor}
\usepackage{symbols}

\usepackage{amsmath,amsfonts,bm}

\def\1{\bm{1}}

\def\sp{space}

\DeclareMathAlphabet{\mathsfit}{\encodingdefault}{\sfdefault}{m}{sl}
\SetMathAlphabet{\mathsfit}{bold}{\encodingdefault}{\sfdefault}{bx}{n}

\newcommand{\R}{\mathbb{R}}

\newcommand{\layer}[1]{\ensuremath{\mathsf{#1}\xspace}}
\usepackage{pifont}
\newcommand{\cmark}{\ding{51}}
\newcommand{\xmark}{\ding{55}}
\usepackage{enumitem}
\usepackage{float}
\setlist{nolistsep}

\definecolor{mybrown}{rgb}{0.87058824, 0.56078431, 0.01960784}
\definecolor{myblue}{rgb}{0.3372549 , 0.70588235, 0.91372549}
\definecolor{mypurple}{rgb}{0.8, 0.47058824, 0.7372549 }
\definecolor{myorange}{rgb}{0.835, 0.368, 0}
\definecolor{mygreen}{rgb}{0.00784314, 0.61960784, 0.45098039}
\definecolor{mygt}{rgb}{0.0078125 , 0.57421875, 0.40625}
\definecolor{mysp}{rgb}{0.84765625, 0.515625  , 0.0234375}
\definecolor{mycitecolor}{rgb}{0,0.08,0.45}
\definecolor{mygr}{rgb}{0.9607,0.9607,0.9607}
\definecolor{myoo}{rgb}{0.992,0.9176,0.9019}
\definecolor{myrr}{HTML}{AE031A}
\definecolor{mybb}{HTML}{0155B3}
\usepackage[pagebackref=true,
            breaklinks=true,
            colorlinks,
            bookmarks=false,
            citecolor=mycitecolor,
            urlcolor=magenta,
            hypertexnames=false]{hyperref}

\begin{document}
\pagestyle{headings}
\mainmatter

\def\ACCV22SubNumber{}
\newcommand{\papertitle}{Inverting Adversarially Robust Networks\\for Image Synthesis}
\title{\papertitle}
\titlerunning{Inverting Adversarially Robust Networks for Image Synthesis}
\authorrunning{R. A. Rojas-Gomez et al.}
\author{Renan A. Rojas-Gomez\inst{1} \and Raymond A. Yeh\inst{2} \and\\Minh N. Do\inst{1} \and Anh Nguyen\inst{3}}
\institute{University of Illinois at Urbana-Champaign\and Purdue University \and Auburn University\ \\\email{\{renanar2,minhdo\}@illinois.edu}\quad \email{rayyeh@purdue.edu}\quad \email{anh.ng8@gmail.com}}

\maketitle
\begin{abstract}
Despite unconditional feature inversion being the foundation of many image synthesis applications, training an inverter demands a high computational budget, large decoding capacity and imposing conditions such as autoregressive priors. To address these limitations, we propose the use of adversarially robust representations as a perceptual primitive for feature inversion. We train an adversarially robust encoder to extract disentangled and perceptually-aligned image representations, making them easily invertible. By training a simple generator with the mirror architecture of the encoder, we achieve superior reconstruction quality and generalization over standard models. Based on this, we propose an adversarially robust autoencoder and demonstrate its improved performance on style transfer, image denoising and anomaly detection tasks. Compared to recent ImageNet feature inversion methods, our model attains improved performance with significantly less complexity.\footnote{Code available at \url{https://github.com/renanrojasg/adv_robust_autoencoder}}
\end{abstract}

\section{Introduction}
\label{sec:introduction}

Deep classifiers trained on large-scale datasets extract meaningful high-level features of natural images, making them an essential tool for manipulation tasks such as style transfer~\cite{gatys_2016_image,li_2017_universal,yoo_2019_photorealistic}, image inpainting~\cite{yang_2017_high,nguyen2017plug}, image composition~\cite{shocher_2020_semantic,nguyen2016synthesizing}, among others \cite{rombach_2020_network,santurkar_2019_image,zhang_2018_unreasonable}. State-of-the-art image manipulation techniques use a decoder \cite{nguyen2017plug,zhang_2018_unreasonable}, \ie, an \emph{image generator}, to create natural images from high-level features. Extensive work has explored how to train image generators, leading to models with photorealistic results \cite{goodfellow2016deep}. Moreover, by learning how to invert deep features, image generators enable impressive synthesis use cases such as anomaly detection \cite{deecke_2018_image,golan_2018_deep} and neural network visualization \cite{nguyen2016synthesizing,nguyen2019understanding,ponce2019evolving,rombach2022invertible}. 

Inverting ImageNet features is a challenging task that often requires the generator to be more complex than the encoder~\cite{bigbigan,dosovitskiy2015inverting,dosovitskiy_2016_generating,nguyen2017plug}, incurring in a high computational cost. Donahue et al.~\cite{bigbigan} explained this shortcoming by the fact that the encoder bottleneck learns entangled representations that are hard to invert. An alternative state-of-the-art technique for inverting ImageNet features requires, in addition to the encoder and decoder CNNs, \emph{an extra} autoregressive model and vector quantization \cite{esser2021taming,esser2021imagebart} or a separate invertible network \cite{rombach2022invertible}.

In this paper, we propose \textbf{a novel mechanism for training effective ImageNet autoencoders that do not require extra decoding layers or networks besides the encoder and its mirror decoder}. Specifically, we adopt a pre-trained classifier as encoder and train an image generator to invert its features, yielding an autoencoder for real data. Unlike existing works that use feature extractors trained on natural images, we train the encoder on adversarial examples~\cite{madry_2018_towards}. This fundamental difference equips our \emph{adversarially robust} (AR) autoencoder with representations that are perceptually-aligned with human vision~\cite{engstrom_2019_adversarial,santurkar_2019_image}, resulting in favorable inversion properties.

To show the advantages of learning how to invert AR features, our generator corresponds to the \emph{mirror} architecture of the encoder, without additional decoding layers \cite{bigbigan,shocher_2020_semantic} or extra components \cite{razavi2019generating,esser2021taming,rombach2022invertible,esser2021imagebart,van2017neural}. To the best of our knowledge, we are the first to show the benefits of training an autoencoder on both adversarial and real images. Our main findings are as follows:

\begin{itemize}
    \item A generator trained to invert AR features has a substantially higher reconstruction quality than those trained on standard features. 
    Our method generalizes to different models (AlexNet~\cite{krizhevsky_2012_imagenet}, VGG-16~\cite{simonyan_2014_very}, and ResNet~\cite{he_2016_deep}) and datasets (CIFAR-10~\cite{krizhevsky_2009_learning} and ImageNet~\cite{russakovsky_2015_imagenet})(\secref{sec:experimental_inverting}).
    \item Our proposed AR autoencoder is remarkably robust to resolution changes, as shown on natural and upscaled high-resolution images (Fig.~\ref{fig:hires}). Experiments on DIV2K \cite{agustsson_2017_ntire} show it accurately reconstructs high-resolution images without any finetuning, despite being trained on low-resolution images (\secref{sec:experimental_scale}).
    \item  Our generator outperforms state-of-the-art inversion methods based on iterative optimization techniques \cite{engstrom_2019_adversarial} in terms of PSNR, SSIM, and LPIPS \cite{zhang_2018_unreasonable}. It also attains comparable accuracy to the well-established DeepSiM model \cite{dosovitskiy_2015_inverting} with a much lower model complexity (\secref{sec:experimental_comparison}).
    \item Our AR model outperforms standard baselines on three downstream tasks: style transfer \cite{li_2017_universal}, image denoising \cite{nguyen2017plug} (Sec.~\ref{sec:tasks}) and anomaly detection  \cite{deecke_2018_image}. The latter is covered in detail in the Appendix (Sec.~\ref{sec:supp_anomaly_detection}).
\end{itemize}

\section{Related Work}
\label{sec:related}

$\quad$
\textbf{Inverting Neural Networks.}
Prior work exploring deep feature inversion using optimization approaches are either limited to per-pixel priors or require multiple steps to converge and are sensitive to initialization \cite{mahendran_2015_understanding,mahendran_2016_visualizing,engstrom_2019_adversarial,santurkar_2019_image}. Instead, we propose to map contracted features to images via a generator, following the work by Dosovitskiy et al. \cite{dosovitskiy_2016_generating} and similar synthesis techniques \cite{shocher_2020_semantic,nguyen2017plug,nguyen2016synthesizing}. By combining natural priors and AR features, we get a significant reconstruction improvement with much less trainable parameters.

Our results are consistent to prior findings on AR features being more invertible via optimization \cite{engstrom_2019_adversarial} and more useful for transfer learning \cite{salman_2020_adversarially}. As part of our contribution, we complement these by showing that (i) learning a map from the AR feature space to the image domain largely outperforms the original optimization approach, (ii) such an improvement generalizes to models of different complexity, and (iii) inverting AR features shows remarkable robustness to scale changes. We also show AR encoders with higher robustness can be more easily decoded, revealing potential security issues \cite{zhang2020secret}.

\textbf{Regularized Autoencoders.}
Prior work requiring data augmentation to train generative and autoencoding models often requires learning an invertible transformation  that maps augmented samples back to real data \cite{jun2020distribution}. Instead, our approach can be seen as a novel way to regularize bottleneck features, providing an alternative to contractive, variational and sparse autoencoders \cite{goodfellow2016deep,kingma_2013_auto,ng2011sparse}.

\section{Preliminaries}
\label{sec:framework}
Our model exploits AR representations to reconstruct high-quality images, which is related to the feature inversion framework. Specifically, we explore AR features as a strong prior to obtain photorealism. For a clear understanding of our proposal, we review fundamental concepts of feature inversion and AR training.

{\bf Feature Inversion.}
Consider a target image $x_{0}\in \mathbb{R}^{W \times H \times C}$ and its contracted representation $f_{0}\triangleq F_{\theta}(x_{0})\in \mathbb{R}^{W'\times H'\times C'}$. Here, $F_{\theta}$ denotes the target model, \eg AlexNet, with parameters $\theta \in \mathbb{R}^{T}$ and $W'H'C'\ll WHC$. Features extracted by $F_{\theta}$ encapsulate rich input information that can either be used for the task it was trained on, transferred to a related domain \cite{pan_2009_survey} or  used for applications such as image enhancement and manipulation \cite{johnson_2016_perceptual,gatys_2016_image}.

An effective way to leverage these representations is by training a second model, a generator, to map them to the pixel domain. This way, deep features can be manipulated and transformed into images \cite{dosovitskiy_2015_inverting,dosovitskiy_2016_generating}. Also, since deep features preserve partial input information, inverting them elucidates what kind of attributes they encode. Based on these, \textit{feature inversion} \cite{simonyan_2013_deep,mahendran_2015_understanding,dosovitskiy_2015_inverting} has been extensively studied for visualization and understanding purposes as well as for synthesis and manipulation tasks. Typically, feature inversion is formulated as an optimization problem:
\begin{align}
    \hat{x}=&\ \text{arg } \underset{x}{\text{min}}\ \mathcal{F}(F_{\theta}(x), f_{0}) + \lambda \mathcal{R}(x),
\end{align}
where $\hat{x}$ is the estimated image and $\mathcal{F}(F_{\theta}(x), f_{0})$ the fidelity term between estimated and target representations, $F_{\theta}(x)$ and $f_{0}$ respectively. $\mathcal{R}(x)$ denotes the regularization term imposing \textit{apriori} constraints in the pixel domain and $\lambda \in \mathbb{R}_{++}$ balances between fidelity and regularization terms.

{\bf Adversarial Robustness.}
Adversarial training adds perturbations to the input data and lets the network learn how to classify in the presence of such adversarial attacks \cite{goodfellow_2014_explaining,madry_2018_towards,athalye_2018_obfuscated}. Consider the image classification task with annotated dataset $\mathcal{K}$. Let an annotated pair correspond to image $x \in \mathbb{R}^{W\times H\times C}$ and its one-hot encoded label $y\in \{0,1\}^{|\mathcal{C}|}$, where $\mathcal{C}$ is the set of possible classes. From the definition by Madry et al.~\cite{madry_2018_towards}, a perturbed input is denoted by $x'=x+\delta$, where $x'$ is the perturbed sample and $\delta$ the perturbation. Let the set of perturbations be bounded by the $\ell_{p}$ ball for $p\in \{2,\infty\},\ \mathcal{S}:\{\delta,\ \|\delta\|_{p}\leq\varepsilon\}$. Then, the AR training corresponds to an optimization problem:
\vspace{-0.1cm}
\begin{align}
    \label{eq:robust01}
    \tilde{\theta}=&\ \text{arg }\underset{\theta}{\text{min}}\ \mathbb{E}_{(x,y)\sim \mathcal{K}}\bigg[\underset{\delta \in \mathcal{S}}{\text{max}}\ \mathcal{L}_{x', y}(\theta) \bigg],
    \vspace{-0.1cm}
\end{align}
where $\tilde{\theta}\in \mathbb{R}^{T}$ are the optimal weights and $\mathcal{L}_{x',y}(\theta)$ the negative log-likelihood. The goal is to minimize $\mathcal{L}_{x',y}(\theta)$ in the presence of the worst possible adversary.

\section{Proposed Method}
\label{sec:proposed}

\subsection{Adversarially Robust Autoencoder}
\label{sec:proposed_inversion}

\begin{figure}[t]
\hspace{0.55\textwidth}\begin{minipage}{0.15\textwidth}
\centering \textbf{\colorbox{white}{\scalebox{.7}{Ground-truth}}}
\end{minipage}\begin{minipage}{0.15\textwidth}
\centering \textbf{\colorbox{white}{\scalebox{.7}{Standard}}}
\end{minipage}\begin{minipage}{0.15\textwidth}
\centering \textbf{\colorbox{white}{\scalebox{.7}{AR (ours)}}}
\end{minipage}

\vspace{-0.9 cm}
\subfloat[Proposed Adversarially Robust Model]{\includegraphics[width=0.45\textwidth]{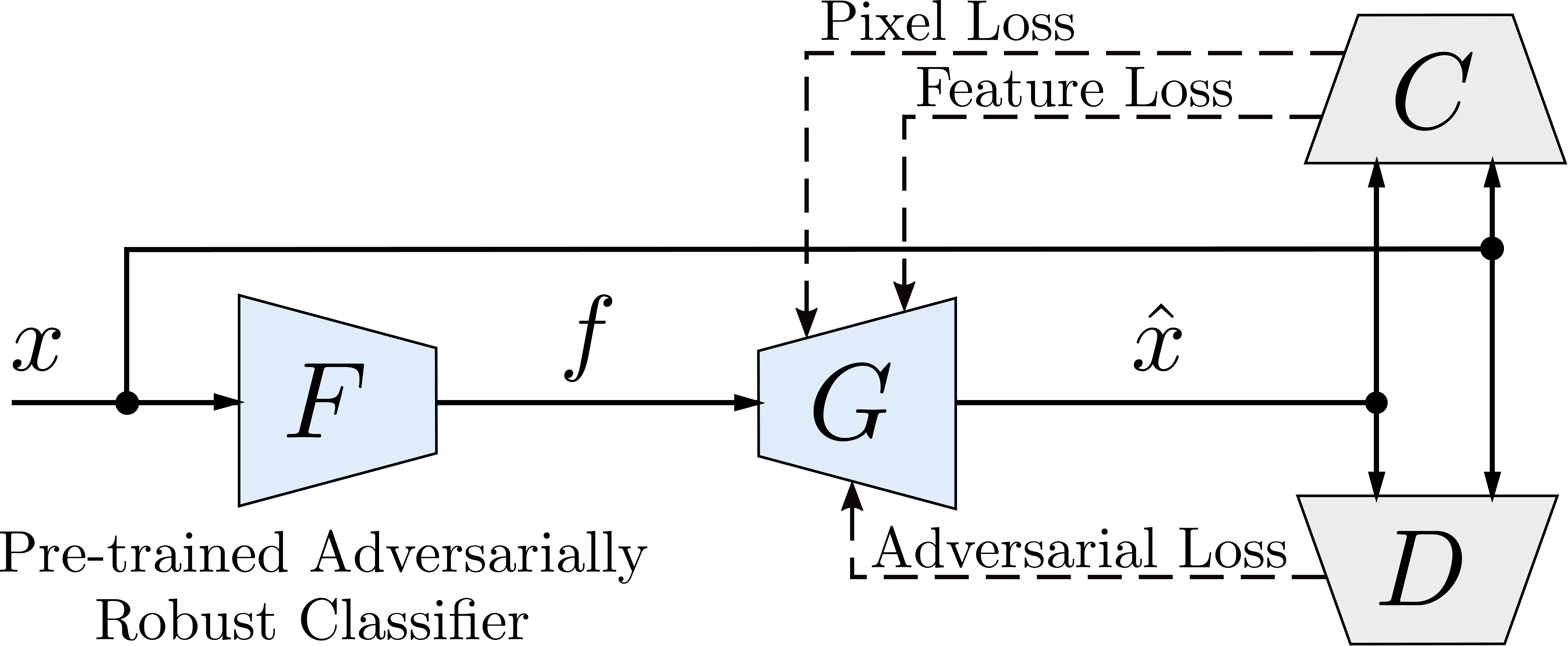}\hspace{0.1\textwidth}}
\subfloat[Feature Inversion ($224 \times 224$ px.)]{\includegraphics[width=0.45\textwidth]{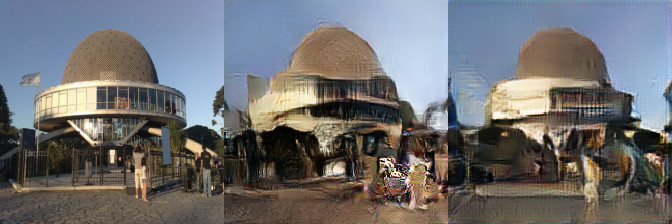}}

\caption{\label{fig:proposed_model}By training it to invert adversarially robust features, our proposed autoencoder obtains better reconstructions than models trained on standard features.}
\vspace{-0.5cm}
\end{figure}

We propose an autoencoder architecture (\figref{fig:proposed_model}) to extract bottleneck AR features of arbitrary input images, manipulate them for a given synthesis task, and map the results back to images. We denote the AR feature extractor as $F_{\tilde{\theta}}$, where $\tilde{\theta}$ are the AR model weights, as explained in \secref{sec:framework}. Robust features are transformed into images using a CNN-based generator denoted as $G_{\tilde{\phi}}$. Here, $\tilde{\phi}$ are the generator weights learned by inverting AR features.

Following prior works~\cite{ulyanov2018deep,dosovitskiy_2016_generating}, we use AlexNet as the encoder and extract AR features from its \layer{conv5} layer. We also explore more complex encoders from the VGG and ResNet families and evaluate their improvement over standard encoders (See \secref{sec:supp_proposed} for architecture details).

\subsection{Image Decoder: Optimization Criteria}
\label{sec:opt_crit}

Given a pre-trained AR encoder $F_{\tilde{\theta}}$, the generator $G_{\tilde{\phi}}$ is trained using $\ell_{1}$ pixel, $\ell_{2}$ feature and GAN losses, where the feature loss matches AR representations, known to be \textit{perceptually aligned} \cite{engstrom_2019_adversarial}.

In more detail, we denote $\hat{x}=G_{\tilde{\phi}}(f)$ to be the reconstruction of image $x$, where $f=F_{\tilde{\theta}}(x)$ are its AR features. Training the generator with fixed encoder's weights $\tilde{\theta}$  corresponds to the following optimization problem:
\begin{align}
    \tilde{\phi}=\ \text{arg }\underset{\phi}{\text{min}}&\ \lambda_{\text{pix}}\mathcal{L}_{\text{pix}}(\phi)+\lambda_{\text{feat}}\mathcal{L}_{\text{feat}}(\phi, \tilde{\theta}) + \lambda_{\text{adv}}\mathcal{L}_{\text{adv}}(\phi,\psi),
\end{align}
\vspace{-1.5\baselineskip}
\begin{align}
    \mathcal{L}_{\text{pix}}(\phi)\triangleq& \ \mathbb{E}_{x\sim \tilde{\mathcal{K}}}\ \|x-G_{\phi}(f)\|_{1},\\
    \mathcal{L}_{\text{feat}}(\phi, \tilde{\theta})\triangleq& \ \mathbb{E}_{x\sim \tilde{\mathcal{K}}}\ \|f-F_{\tilde{\theta}}\circ G_{\phi}(f)\|_{2}^{2},\\
    \mathcal{L}_{\text{adv}}(\phi, \psi)\triangleq&\ \mathbb{E}_{x\sim \tilde{\mathcal{K}}}\big[-\text{log}D_{\psi}\circ G_{\phi}(f)\big],
\end{align}
where $\lambda_{\text{pix}}, \lambda_{\text{feat}}, \lambda_{\text{adv}}\in \R_{++}$ are hyperparameters,  $D_{\psi}: \R^{W \times H \times C}\mapsto [0,1]$ denotes the discriminator with weights $\psi$ and predicts the probability of an image being real.
The pixel loss $\mathcal{L}_{\text{pix}}(\phi)$ is the $\ell_{1}$ distance between prediction $G_{\phi}(f)$ and target $x$. The feature loss $\mathcal{L}_{\text{feat}}(\phi, \theta)$ is the $\ell_{2}$ distance between the AR features of prediction and target. The adversarial loss $\mathcal{L}_{\text{adv}}(\phi, \psi)$ maximizes the discriminator score of predictions, \ie, it increases the chance the discriminator classifies them as real. On the other hand, the discriminator weights are trained via the cross-entropy loss, \ie, 
\begin{align}
 \min_\psi \mathcal{L}_{\text{disc}}(\phi, \psi)\triangleq&\ \mathbb{E}_{x\sim \tilde{\mathcal{K}}}\big[-\text{log}D_{\psi}(x)-\text{log}(1-D_{\psi}\circ G_{\phi}(f))\big].
\end{align}
This discriminative loss $ \mathcal{L}_{\text{disc}}(\phi, \psi)$ guides $D_\psi$ to maximize the score of real images and minimize the score of reconstructed (fake) images. Similar to traditional GAN algorithms, we alternate between the generator and discriminator training to reach the equilibrium point.

\subsection{Applications}
\label{sec:applications}

The trained AR autoencoder can be used to improve the performance of tasks such as style transfer \cite{li_2017_universal}, image denoising \cite{vincent_2010_stacked}, and anomaly detection \cite{deecke_2018_image}. In what follows, we describe the use of our model on style transfer and image denoising. The task of anomaly detection is covered in the Appendix (\secref{sec:supp_anomaly_detection}).

\begin{figure}[t]
\hspace{0.5\textwidth}
\begin{minipage}{0.09\textwidth}
\centering\textbf{\scriptsize{Refs}}
\end{minipage}\begin{minipage}{0.225\textwidth}
\centering\textbf{\scriptsize{Standard}}
\end{minipage}\begin{minipage}{0.175\textwidth}
\centering\textbf{\scriptsize{AR (ours)}}
\end{minipage}\vspace{-2\baselineskip}

\subfloat[Multilevel Stylization]{\includegraphics[width=0.375\textwidth]{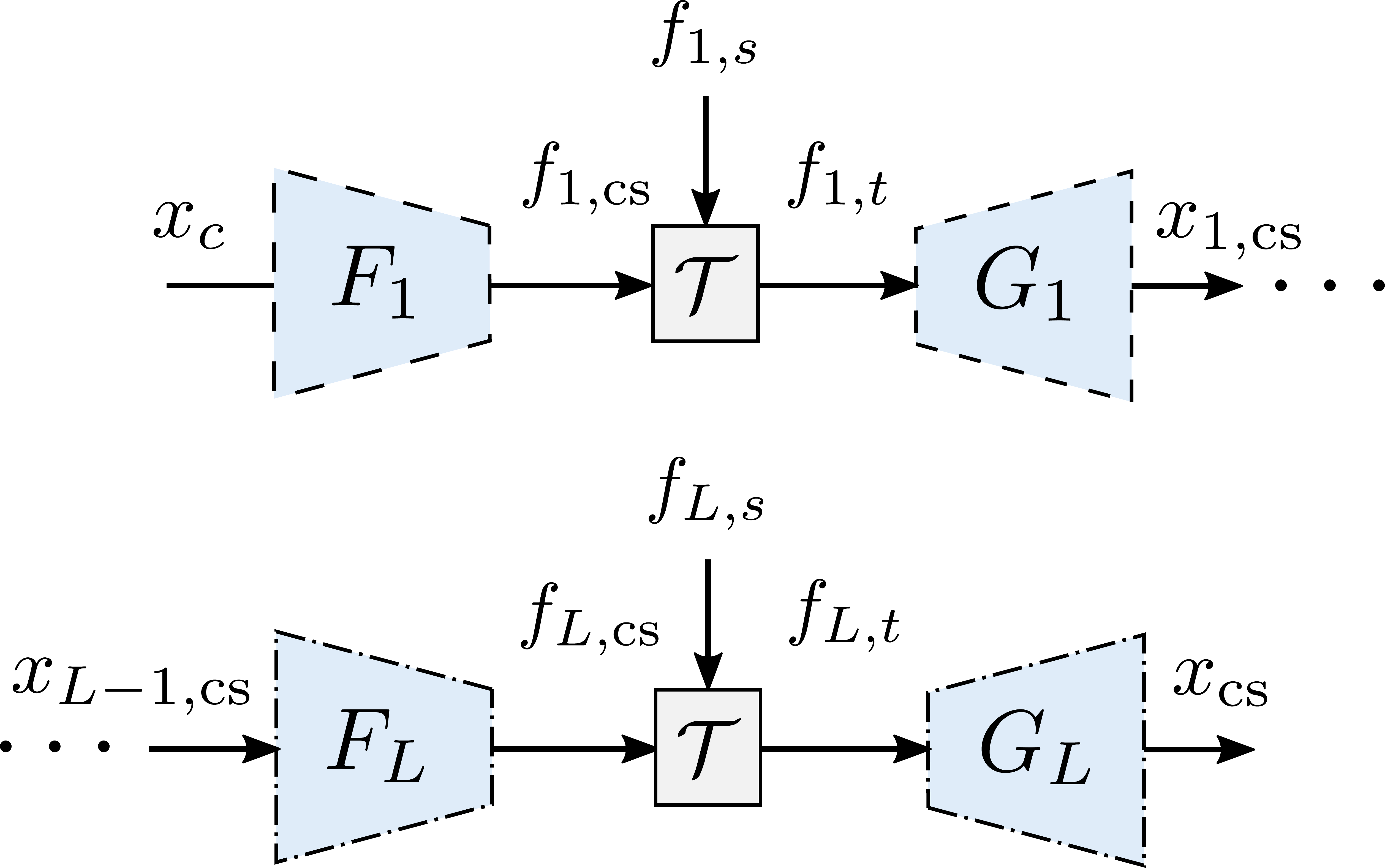}}\hspace{0.125\textwidth}
\subfloat[Adversarially Robust Style Transfer]{\includegraphics[width=0.5\textwidth]{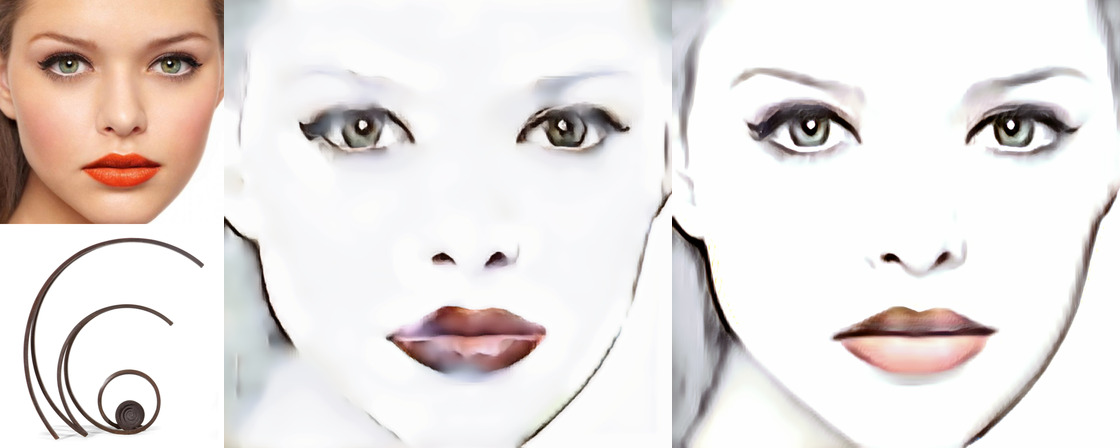}}

\vspace{-0.2cm}
\caption{\label{fig:proposed_st}Example-based Style Transfer using adversarially robust features.}
\vspace{-0.2cm}
\end{figure}

\textbf{Example-based Style Transfer.} Style transfer \cite{gatys_2016_image} aligns deep features to impose perceptual properties of a style image $x_{s}$ over semantic properties of a content image $x_{c}$. This is done by matching the content and style distributions in the latent space of a pre-trained encoder to then transform the resulting features back into images. We adopt the Universal Style Transfer framework \cite{li_2017_universal} to show the benefits of using our AR model for stylization (\figref{fig:proposed_st}).

\begin{figure}[t]

\hspace{0.45\textwidth}
\begin{minipage}[t]{0.1375\textwidth}
\centering\textbf{\colorbox{white}{\scalebox{.7}{\hspace{-0.35\baselineskip}Ground-truth}}}
\end{minipage}\begin{minipage}[t]{0.1375\textwidth}
\centering \textbf{\colorbox{white}{\scalebox{.7}{Observation}}}
\end{minipage}\begin{minipage}[t]{0.1375\textwidth}
\centering \textbf{\colorbox{white}{\scalebox{.7}{Standard}}}
\end{minipage}\begin{minipage}[t]{0.1375\textwidth}
\centering \textbf{\colorbox{white}{\scalebox{.7}{AR (ours)}}}

\end{minipage}\vspace{-0.5 cm}
\subfloat[Skip connected AR model]{\includegraphics[width=0.4\textwidth]{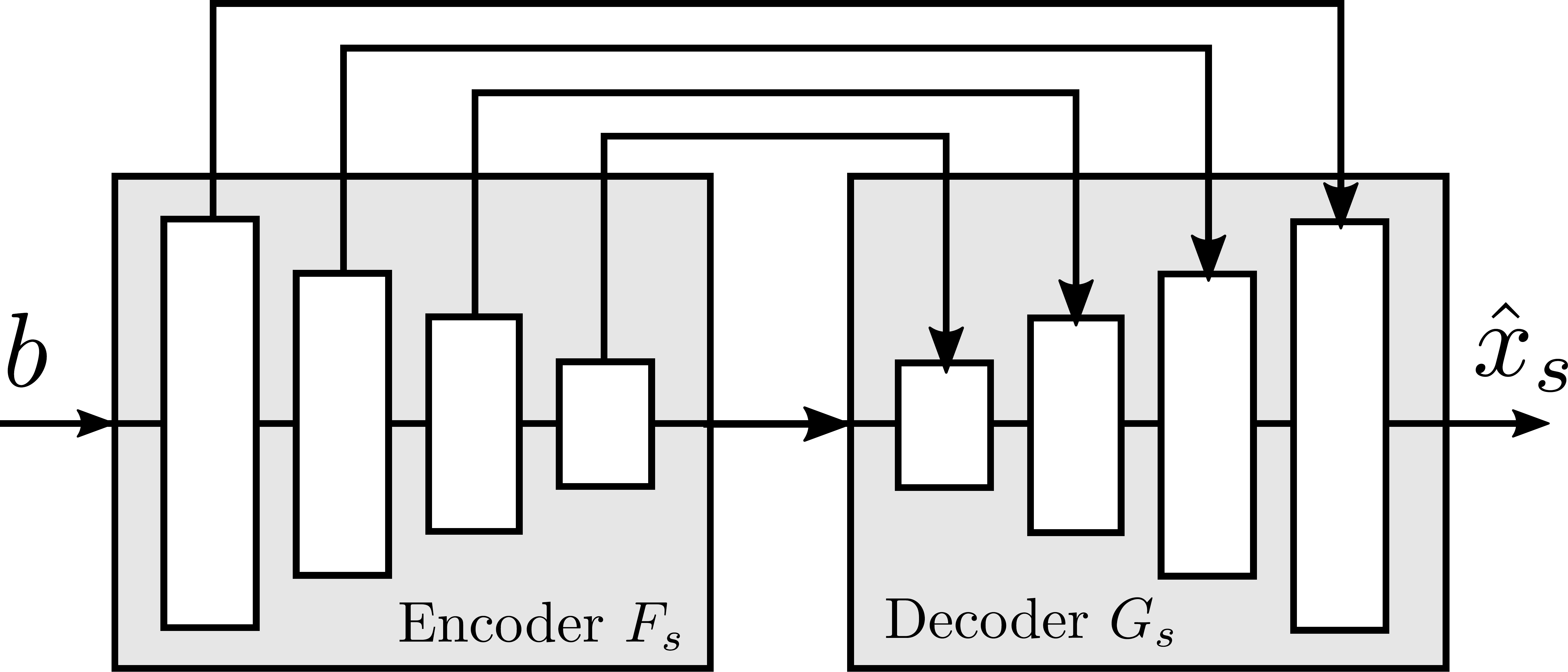}}\hspace{0.05\textwidth}
\subfloat[Adversarially Robust Image Denoising]{\includegraphics[width=0.55\textwidth]{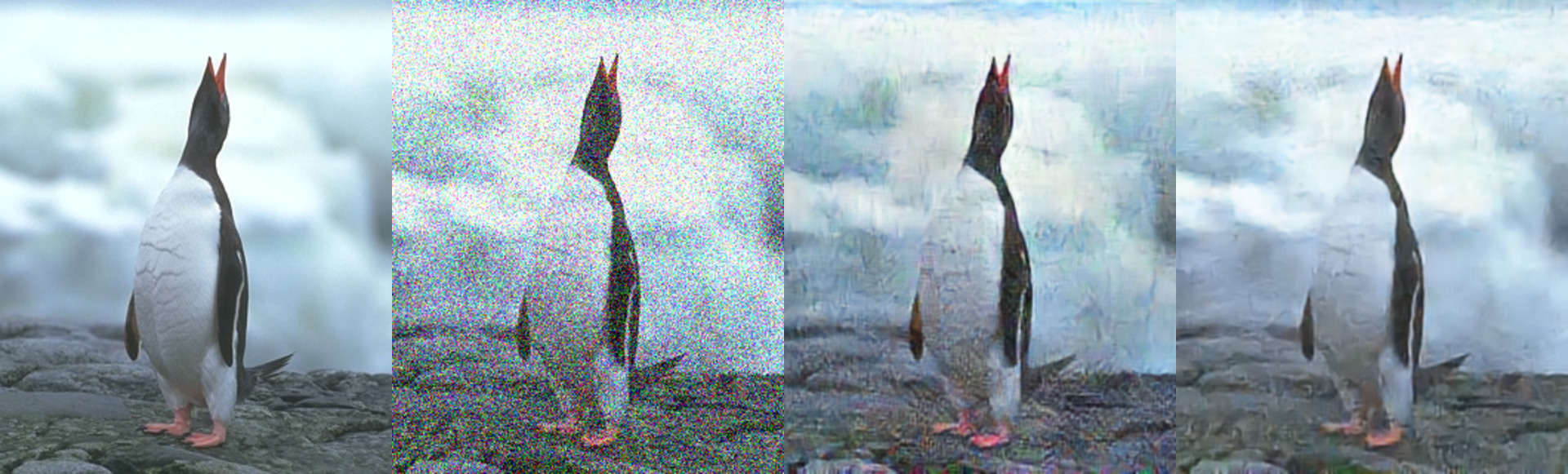}}

\vspace{-0.2 cm}
\caption{\label{fig:proposed_denoising}Image denoising using our adversarially robust autoencoder.}
\vspace{-0.7 cm}
\end{figure}

We train three AR AlexNet autoencoders $\{F_{l,\tilde{\theta}}, G_{l,\tilde{\phi}}\}_{l=1}^{L=3}$ and use them to sequentially align features at each scale. $F_{1,\tilde{\theta}}$, $F_{2,\tilde{\theta}}$ and $F_{3,\tilde{\theta}}$ extract AR \layer{conv5}, \layer{conv2} and \layer{conv1} features, respectively. First, style features $f_{l,s}=F_{l,\tilde{\theta}}(x_{s})$ are extracted at each stage. We then use the content image as initialization for the stylized output $x_{0,\text{cs}}\triangleq x_{c}$ and extract its \layer{conv5} features $f_{1,\text{cs}}=F_{1,\tilde{\theta}}(x_{0,\text{cs}})$.

At stage $l=1$, the style distribution is imposed over the content features by using the whitening and coloring transform \cite{li_2017_universal,kessy_2018_optimal} denoted by $\mathcal{T}$. The resulting representation $f_{1,t}= \mathcal{T}\big(f_{1,s},f_{1,\text{cs}}\big)$ is characterized by the first and second moments of the style distribution. An intermediate stylized image $x_{1,\text{cs}}$ incorporating the style at the first scale is then generated as $x_{1,\text{cs}}= G_{1,\tilde{\phi}}(f_{1,t})$.

The process is repeated for $l\in \{2,3\}$ to incorporate the style at finer resolutions, resulting in the final stylized image $x_{\text{cs}}= x_{3,\text{cs}}$.

\textbf{Image Denoising.} Motivated by denoising autoencoders (DAE) \cite{vincent_2010_stacked} where meaningful features are extracted from distorted instances, we leverage AR features for image enhancement tasks. Similarly to deep denoising models \cite{mao_2016_image}, we incorporate skip connections in our pre-trained AR AlexNet autoencoder to extract features at different scales, complementing the information distilled at the encoder bottleneck (\figref{fig:proposed_denoising}).
Skip connections correspond to Wavelet Pooling \cite{yoo_2019_photorealistic}, replacing pooling and upsampling layers by analysis and synthesis Haar wavelet operators, respectively. Our skip-connected model is denoted by $\{F_{s, \tilde{\theta}}, G_{s, \tilde{\phi}}\}$.

Similarly to real quantization scenarios \cite{el_2020_blind,zhang_2018_ffdnet,moeller_2015_learning}, we assume images are corrupted by clipped additive Gaussian noise. A noisy image is denoted by $b= \rho\big(x+\eta\big)\ \in \R^{W\times H\times C}$, where $\eta\sim \mathcal{N}(0, \sigma)$ is the additive white Gaussian noise term and $\rho(x)= \max[0, \min(1, x)]$ a pointwise operator restricting the range between $0$ and $1$. Denoised images are denoted by $\hat{x}_{s}= G_{s,\tilde{\phi}}\circ F_{s,\tilde{\theta}}(b)$.

$G_{s, \tilde{\phi}}$ is trained to recover an image $x$ from the features of its corrupted version $F_{s, \tilde{\theta}}(b)$. The training process uses the optimization criteria described in \secref{sec:opt_crit}.

\section{Experiments on Feature Inversion}
\label{sec:experimental}

\begin{table}[t]
\vspace{0.3 cm}
\small
\def\arraystretch{1.25}
\setlength\tabcolsep{4pt}
\begin{center}
\fontsize{7}{7}\selectfont
\caption{\label{tab:inversion_alexnet}AlexNet feature inversion on ImageNet. Under distinct training losses, inverting AR features via our proposed generator is consistently more accurate than inverting standard features.}
\vspace{0.1cm}
\begin{tabular}{c|c|c|c|c} 
\specialrule{.15em}{.05em}{.05em} 
Losses & Model & PSNR (dB)$\uparrow$ & SSIM$\uparrow$ & LPIPS$\downarrow$\\
\hline
 \multirow{2}{*}{\makecell{Pixel}} & \makecell{Standard} & $17.562\pm 2.564$ & $0.454\pm 0.167$ & $0.624\pm 0.099$\\
 & \makecell{AR (ours)} & $\mathbf{19.904\pm 2.892}$ & $\mathbf{0.505\pm 0.169}$ & $\mathbf{0.596\pm 0.104}$\\
 \hline
 \multirow{2}{*}{\makecell{Pixel, Feature}} & \makecell{Standard} & $14.462\pm 1.884$ & $0.103\pm 0.044$ & $0.713\pm 0.046$\\
 & \makecell{AR (ours)} & $\mathbf{17.182\pm 2.661}$ & $\mathbf{0.284\pm 0.111}$ & $\mathbf{0.601\pm 0.034}$\\  
 \hline
 \multirow{2}{*}{\makecell{Pixel, Feature,\\GAN}} & \makecell{Standard} & $15.057\pm 2.392$ & $0.307 \pm 0.158$ & $\mathbf{0.547\pm 0.055}$\\
 & \makecell{AR (ours)} & $\mathbf{17.227\pm 2.725}$ & $\mathbf{0.358\pm 0.163}$ & $0.567\pm 0.056$\\
\specialrule{.15em}{.05em}{.05em} 
\end{tabular}
\end{center}
\vspace{-0.4 cm}
\end{table}

\begin{figure}[t]
\centering
\begin{minipage}[t]{0.1429\textwidth}
\centering \colorbox{white}{\textbf{\scalebox{.8}{G. truth}}}
\end{minipage}\begin{minipage}[t]{0.1429\textwidth}
\centering \colorbox{white}{\textbf{\scalebox{.8}{Standard}}}
\end{minipage}\begin{minipage}[t]{0.1429\textwidth}
\centering \colorbox{white}{\textbf{\scalebox{.8}{AR (ours)}}}
\end{minipage}\begin{minipage}[t]{0.1429\textwidth}
\centering \colorbox{white}{\textbf{\scalebox{.8}{Standard}}}
\end{minipage}\begin{minipage}[t]{0.1428\textwidth}
\centering \colorbox{white}{\textbf{\scalebox{.8}{AR (ours)}}}
\end{minipage}\begin{minipage}[t]{0.1429\textwidth}
\centering \colorbox{white}{\textbf{\scalebox{.8}{Standard}}}
\end{minipage}\begin{minipage}[t]{0.1429\textwidth}
\centering \colorbox{white}{\textbf{\scalebox{.8}{AR (ours)}}}
\end{minipage}\vspace{-0.4\baselineskip}

\begin{minipage}{0.1429\textwidth}
\centering\textbf{ }
\end{minipage}\begin{minipage}[t]{0.1429\textwidth}
\centering \scalebox{0.65}{Pix.}
\end{minipage}\begin{minipage}[t]{0.1429\textwidth}
\centering \scalebox{0.65}{Pix.}
\end{minipage}\begin{minipage}[t]{0.1429\textwidth}
\centering \scalebox{0.65}{Pix., Feat.}
\end{minipage}\begin{minipage}[t]{0.1428\textwidth}
\centering \scalebox{0.65}{Pix., Feat.}
\end{minipage}\begin{minipage}[t]{0.1429\textwidth}
\centering \scalebox{0.65}{Pix., Feat., GAN}
\end{minipage}\begin{minipage}[t]{0.1429\textwidth}
\centering \scalebox{0.65}{Pix., Feat., GAN}
\end{minipage}

\includegraphics[width=\textwidth]{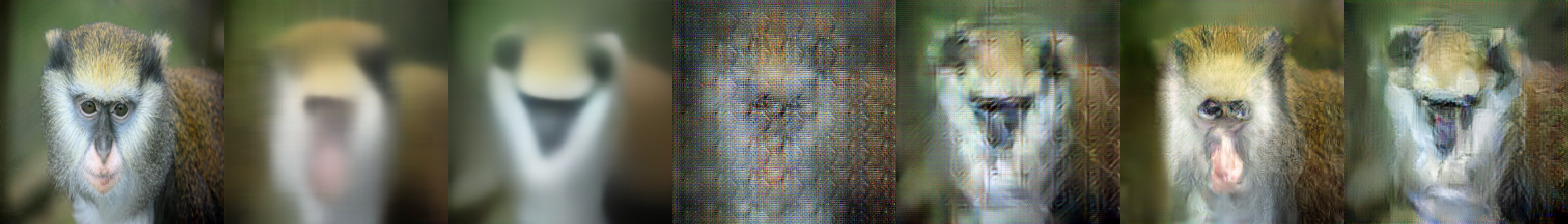}

\includegraphics[width=\textwidth]{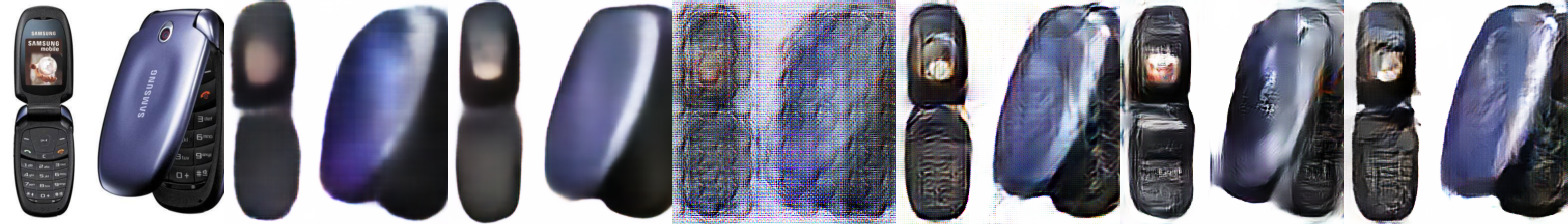}

\caption{\label{fig:inversion_alexnet}AlexNet feature inversion on ImageNet. \layer{Conv5} features are inverted using our proposed generator under three different training criteria. Reconstructions from AR features are more faithful to the ground-truth image.}
\vspace{-0.4cm}
\end{figure}

We begin analyzing the reconstruction accuracy achieved by inverting features from different classifiers and empirically show that learning how to invert AR features via our proposed generator improves over standard feature inversion. Refer to \secref{sec:supp_results} and \secref{sec:supp_proposed_method} for additional inversion results and training details.

\subsection{Reconstruction Accuracy of AR Autoencoders}
\label{sec:experimental_inverting}
\textbf{Inverting AlexNet features.} Standard and AR AlexNet autoencoders are trained as described in \secref{sec:proposed_inversion} on ImageNet for comparison purposes. The AR AlexNet classifier is trained via $\ell_{2}$-PGD attacks \cite{madry_2018_towards} of radius $\varepsilon=\frac{3}{255}$ and $7$ steps of size $0.5$. Training is performed using $90$ epochs via SGD with a learning rate of $0.1$ reduced $10$ times every $30$ epochs. On the other hand, the standard AlexNet classifier is trained on natural images via cross-entropy (CE) loss with the same SGD setup as in the AR case.

Next, generators are trained  using pixel, feature and GAN losses to invert AlexNet \layer{conv5} features (size $6\times6\times256$). Both AR and standard models use the same generator architecture, which corresponds to the mirror network of the encoder. We deliberately use a simple architecture to highlight the reconstruction improvement is due to inverting AR features and not the generator capacity. We also train generators using (i) pixel and (ii) pixel and feature losses to ablate their effect. Reconstruction quality is evaluated using PSNR, SSIM and LPIPS.

Under all three loss combinations, reconstructions from AR AlexNet features obtain better PSNR and SSIM than their standard counterparts (\tabref{tab:inversion_alexnet}). Specifically, inverting AR AlexNet features gives an average PSNR improvement of over $2$ dB in all three cases. LPIPS scores also improve, except when using pixel, feature and GAN losses. Nevertheless, inverting AR features obtain a strong PSNR and SSIM improvement in this case as well. Qualitatively, inverting AR features better preserves the natural appearance in all cases, reducing the checkerboard effect and retaining sharp edges (\figref{fig:inversion_alexnet}).

\textbf{Inverting VGG features.} We extend the analysis to VGG-16 trained on ImageNet-143 and evaluate the reconstruction improvement achieved by inverting its AR features. We use the AR pre-trained classifier from the recent work by Liu et al.~\cite{liu_2018_adv} trained using $\ell_{\infty}$-PGD attacks of radius $\varepsilon=0.01$ and $10$ steps of size $\frac{1}{50}$. Training is performed using $80$ epochs via SGD with a learning rate of $0.1$ reduced $10$ times every $30$, $20$, $20$ and $10$ epochs. On the other hand, its standard version is trained on natural images via CE loss with the same SGD setup as in the AR case.

Generators are trained on pixel and feature losses to invert VGG-16 \layer{conv5\_1} features (size $14 \times 14 \times 512$). Similarly to the AlexNet analysis, generators inverting both standard and AR features correspond to the mirror network of the encoder. We evaluate the reconstruction accuracy of both models and report their level of adversarial robustness (\tabref{tab:inversion_vgg16} and \figref{fig:inversion_vgg16}).

\begin{table}[t]

\centering
\begin{minipage}{0.575\textwidth}
\fontsize{8.45}{10.45}\selectfont
\begin{center}
\vspace{-0.8 cm}
\caption{\label{tab:inversion_vgg16} AR VGG-16 \cite{liu_2018_adv} feature inversion on ImageNet. Training our generator via pixel and feature losses, reconstruction largely improves by inverting AR representations.}
\begin{tabular}{c|c|c}
\specialrule{.15em}{.05em}{.05em} 
 & Standard Model & AR Model (ours)\\
 \hline
\makecell{Standard Accuracy} & $65.0$ & $48.7$\\
\makecell{$\ell_{\infty}$ PGD Accuracy} & $0$ & $23.0$\\
\hline
\makecell{PSNR (dB) $\uparrow$} & $18.35\pm 2.471$ & $\mathbf{21.063\pm 3.132}$\\
\makecell{SSIM $\uparrow$} & $0.466\pm 0.2$ & $\mathbf{0.538\pm 0.165}$\\
\makecell{LPIPS $\downarrow$} & $0.327\pm 0.101$ & $\mathbf{0.225\pm0.057}$\\
\specialrule{.15em}{.05em}{.05em} 
\end{tabular}
\end{center}
\end{minipage}
\hfill
\begin{minipage}{0.375\textwidth}
\hspace{2.21\baselineskip}\noindent\fcolorbox{white}{white}{\begin{minipage}[t]{0.155\textwidth}
\centering\textbf{{\scalebox{0.525}{\hspace{-0.25\baselineskip}G. truth}}}
\end{minipage}}\noindent\fcolorbox{white}{white}{\begin{minipage}[t]{0.16\textwidth}
\centering\textbf{{\scalebox{0.525}{\hspace{-0.15\baselineskip}Standard}}}
\end{minipage}}\noindent\fcolorbox{white}{white}{\begin{minipage}[t]{0.15\textwidth}
\centering\textbf{{\scalebox{0.525}{\hspace{-0.55\baselineskip}AR (Ours)}}}
\end{minipage}}

\begin{center}
\vspace{-0.475cm}
\includegraphics[width=0.63\textwidth]{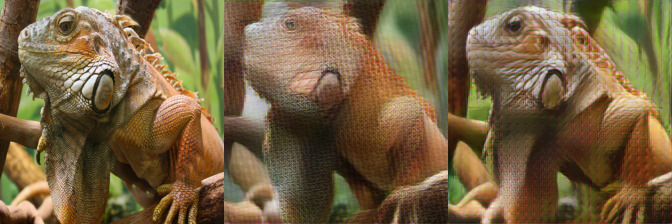}

\includegraphics[width=0.63\textwidth]{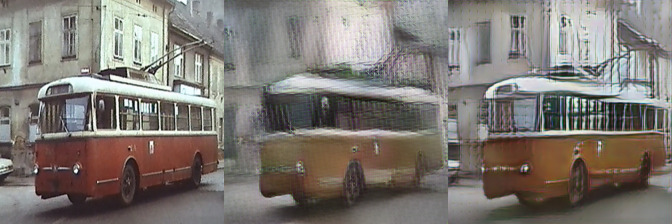}

\includegraphics[width=0.63\textwidth]{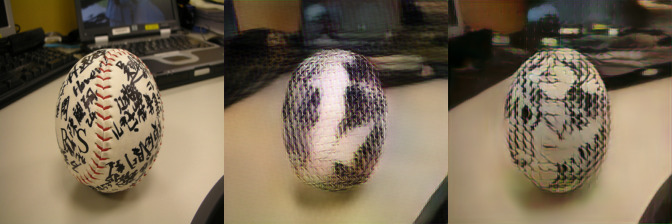}
\end{center}
\captionof{figure}{\label{fig:inversion_vgg16} AR VGG-16 reconstruction on ImageNet.}
\end{minipage}
\vspace{-0.8 cm}
\end{table}

Quantitatively, reconstructions from AR VGG-16 features are more accurate than those of standard features in PSNR, SSIM and LPIPS by a large margin. Specifically, inverting AR VGG-16 features gives an average PSNR improvement of $2.7$ dB. Qualitatively, reconstructions from AR VGG-16 features are more similar to the original images, reducing artifacts and preserving object boundaries.

Furthermore, the reconstruction accuracy attained by the AR VGG-16 autoencoder improves over that of the AR AlexNet model. This suggests that the benefits of inverting AR features are not constrained to shallow models such as AlexNet, but generalize to models with larger capacity.

\textbf{Inverting ResNet features.} To analyze the effect of inverting AR features from classifiers trained on different datasets, we evaluate the reconstruction accuracy obtained by inverting WideResNet-28-10 trained on CIFAR-10. We use the AR pre-trained classifier from the recent work by Zhang et al.~\cite{zhang_2020_geometry}. This model obtains State-of-the-art AR classification accuracy via a novel weighted adversarial training regime. Specifically, the model is adversarially trained via PGD by ranking the importance of each sample based on how close it is to the decision boundary (how \textit{attackable} the sample is).

AR training is performed using $\ell_{\infty}$ attacks of radius $\varepsilon=\frac{8}{255}$ and $10$ steps of size $\frac{2}{255}$. Classification training is performed using $100$ epochs (with a burn-in period of $30$ epochs) via SGD with a learning rate of $0.1$ reduced $10$ times every $30$ epochs. On the other hand, its standard version is trained on natural images via CE loss using the same SGD setup as in the AR case.

Generators for standard and AR WideResNet-28-10 models are trained to invert features from its 3rd residual block (size $8\times 8 \times 640$) via pixel and feature losses. Similarly to our previous analysis, both generators correspond to the mirror architecture of the encoder. We evaluate their reconstruction via PSNR, SSIM and LPIPS, and their robustness via AutoAttack \cite{croce_2020_reliable} (\tabref{tab:inversion_resnet28} and \figref{fig:inversion_resnet28}).

Similarly to previous scenarios, inverting WideResNet-28-10 AR features shows a large improvement over standard ones in all metrics. Specifically, inverting AR features increases PSNR in $4.8$ dB on average over standard features. Visually, the AR WideResNet-28-10 autoencoder reduces bogus components and preserves object contours on CIFAR-10 test samples.

Overall, results enforce our claim that the \textbf{benefits of inverting AR features extend to different models, datasets and training strategies}.
\begin{table}[t]
\vspace{-0.25cm}
\centering
\begin{minipage}{0.575\textwidth}
\fontsize{8.5}{10.5}\selectfont
\begin{center}
\vspace{-0.5 cm}
\caption{\label{tab:inversion_resnet28} AR WideResNet-28-10 \cite{zhang_2020_geometry} feature inversion on CIFAR-10. Inverting AR features via our generator trained on pixel and feature losses significantly improves reconstruction.}
\begin{tabular}{c|c|c}
\specialrule{.15em}{.05em}{.05em} 
 & Standard Model & AR Model (ours)\\
\hline
\makecell{Standard Accuracy} & $93.8$ & $89.36$\\
\makecell{AutoAttack \cite{croce_2020_reliable}} & \makecell{$0$} & \makecell{$59.64$}\\
\hline
PSNR (dB) $\uparrow$ & $17.38\pm 2.039$ & $\mathbf{22.14\pm 1.626}$\\
SSIM $\uparrow$ & $0.59\pm 0.1$ & $\mathbf{0.81\pm 0.067}$ \\
LPIPS $\downarrow$ & $0.2547\pm 0.055$ & $\mathbf{0.2318\pm 0.0833}$\\
  
\specialrule{.15em}{.05em}{.05em} 
\end{tabular}
\end{center}
\end{minipage}
\hfill
\begin{minipage}{0.375\textwidth}
\vspace{0.2 cm}

\hspace{2.21\baselineskip}\noindent\fcolorbox{white}{white}{\begin{minipage}[t]{0.155\textwidth}
\centering\textbf{{\scalebox{0.525}{\hspace{-0.25\baselineskip}G. truth}}}
\end{minipage}}\noindent\fcolorbox{white}{white}{\begin{minipage}[t]{0.16\textwidth}
\centering\textbf{{\scalebox{0.525}{\hspace{-0.15\baselineskip}Standard}}}
\end{minipage}}\noindent\fcolorbox{white}{white}{\begin{minipage}[t]{0.15\textwidth}
\centering\textbf{{\scalebox{0.525}{\hspace{-0.55\baselineskip}AR (Ours)}}}
\end{minipage}}

\begin{center}
\vspace{-0.475cm}
\includegraphics[width=0.63\textwidth]{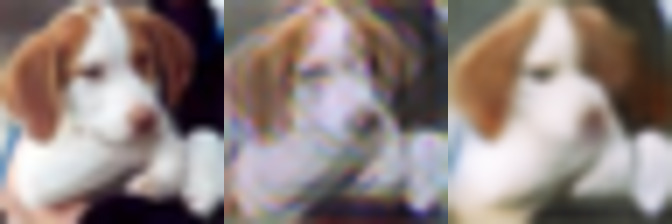}

\includegraphics[width=0.63\textwidth]{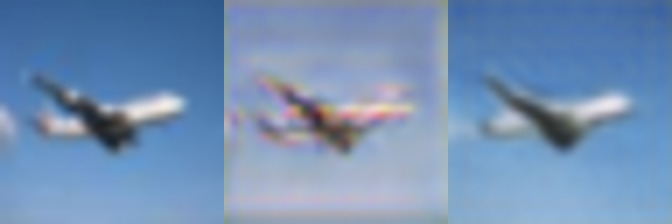}

\includegraphics[width=0.63\textwidth]{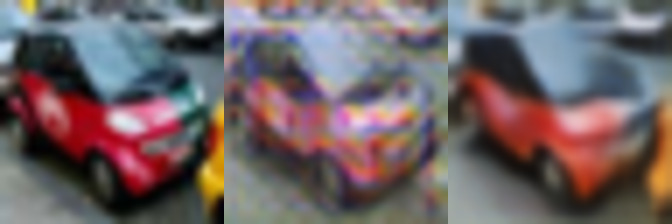}
\end{center}
\captionof{figure}{\label{fig:inversion_resnet28} AR WideResNet-28-10 reconstruction on CIFAR-10.}
\end{minipage}
\vspace{-1 cm}
\end{table}

\subsection{Robustness Level vs. Reconstruction Accuracy}
\label{relationship_adversarial}
We complement the reconstruction analysis by exploring the relation between adversarial robustness and inversion quality. We train five AlexNet classifiers on ImageNet, one on natural images (standard) and four via $\ell_{2}$-PGD attacks with $\varepsilon\in \{0.5,2,3,4\}/255$. All other training parameters are identical across models.

For each classifier, an image generator is trained on an ImageNet subset via pixel, feature and GAN losses to invert \layer{conv5} features. Similar to \secref{sec:experimental_inverting}, all five generators correspond to the mirror network of the encoder. To realiably measure the impact of adversarial robustness, reconstruction accuracy is evaluated in terms of PSNR, SSIM and LPIPS. We also report the effective robustness level achieved by each model via AutoAttack (\tabref{tab:rob_vs_acc}).
\begin{table}[t]
\begin{center}
\vspace{0.3cm}
\caption{\label{tab:rob_vs_acc} Reconstruction vs. Robustness. Experiments on ImageNet show that learning to invert AlexNet features with different AR levels can significantly improve the reconstruction accuracy.}
\resizebox{0.75\columnwidth}{!}{
\begin{tabular}{c|c|c|c|c|c}
\specialrule{.15em}{.05em}{.05em}
& \multicolumn{5}{c}{$\ell_{2}$ PGD Attack ($\varepsilon$)}\\
\cline{2-6}
& $0$ & $0.5$ & $2$ & $3$ & $4$ \\
\hline
\makecell{Standard Accuracy} & $53.69$ & $49.9$ & $43.8$ & $39.83$ & $36.31$\\
\makecell{AutoAttack \cite{croce_2020_reliable}} & \makecell{$8.19$ ($\varepsilon=0.5$)} & \makecell{$48.0$ ($\varepsilon=0.5$)} & \makecell{$28.0$ ($\varepsilon=2$)} & \makecell{$22.27$ ($\varepsilon=3$)} & \makecell{$14.9$ ($\varepsilon=4$)}\\
\hline
PSNR (dB) $\uparrow$ & $13.12$ & $14.41$ & $15.5$ & ${15.53}$ & $\mathbf{15.61}$\\
SSIM $\uparrow$ & $0.20$ & $0.26$ & $\mathbf{0.3}$ & ${0.26}$ & $0.25$\\
LPIPS $\downarrow$ & $0.657$ & ${0.625}$ & $\mathbf{0.614}$ & $0.629$ & $0.644$\\
\specialrule{.15em}{.05em}{.05em} 
\end{tabular}}
\end{center}
\vspace{-0.5cm}
\end{table}

\begin{table}[t]
\centering
\setlength\tabcolsep{2pt}
\begin{minipage}{0.5\textwidth}
\begin{center}
\vspace{-0.8 cm}
\caption{\label{tab:eval_scale} Reconstructing upscaled ImageNet samples. Images upscaled by a factor $L$ are reconstructed from their standard and AR AlexNet features. In contrast to the degraded standard reconstructions, AR reconstructions show an outstanding accuracy that improves for large scaling factors.}
\vspace{0.1 cm}
\fontsize{8.5}{10.5}\selectfont
\def\arraystretch{1.25}
\resizebox{1\columnwidth}{!}{%
\begin{tabular}{c|c|c|c|c|c|c}
\specialrule{.15em}{.05em}{.05em} 
\multirow{2}{*}{\makecell{$L$}} & \multicolumn{3}{c|}{\makecell{Standard\\AlexNet}} & \multicolumn{3}{c}{\makecell{Robust\\AlexNet}} \\
\cline{2-7}
& \makecell{PSNR\\(dB)$\uparrow$} & SSIM$\uparrow$ & LPIPS$\downarrow$ & \makecell{PSNR\\(dB)$\uparrow$} & SSIM$\uparrow$ & LPIPS$\downarrow$\\
\hline
 $1$ & \makecell{$15.057$} & \makecell{$0.3067$} & \makecell{$0.5473$} & \makecell{$17.2273$} & \makecell{$0.3580$} & \makecell{$0.5665$}\\
 $4$ & \makecell{$15.4258$} & \makecell{$0.4655$} & \makecell{$0.4136$} & \makecell{$22.575$} & \makecell{$0.5892$} & \makecell{$0.4012$}\\
 $7$ & \makecell{$13.8922$} & \makecell{$0.4852$} & \makecell{$0.4587$} & \makecell{$23.5778$} & \makecell{$0.6588$} & \makecell{$0.3898$}\\
 $10$ & \makecell{$13.1013$} & \makecell{$0.4969$} & \makecell{$0.486$} & \makecell{$23.9566$} & \makecell{$0.7244$} & \makecell{$0.3892$}\\
\specialrule{.15em}{.05em}{.05em} 
\end{tabular}
}
\end{center}
\end{minipage}
\hfill
\begin{minipage}{0.475\textwidth}

\begin{minipage}{0.2\textwidth}
\centering\textbf{\colorbox{white}{\scalebox{0.68}{G. truth}}}
\end{minipage}\begin{minipage}{0.2\textwidth}
\centering \colorbox{white}{\scalebox{0.7}{$L= 1$}}
\end{minipage}\begin{minipage}{0.2\textwidth}
\centering \colorbox{white}{\scalebox{0.7}{$L= 4$}}
\end{minipage}\begin{minipage}{0.2\textwidth}
\centering \colorbox{white}{\scalebox{0.7}{$L= 7$}}
\end{minipage}\begin{minipage}{0.2\textwidth}
\centering \colorbox{white}{\scalebox{0.7}{$L= 10$}}
\end{minipage}

\vspace{-0.05 cm}
\includegraphics[width=1\textwidth]{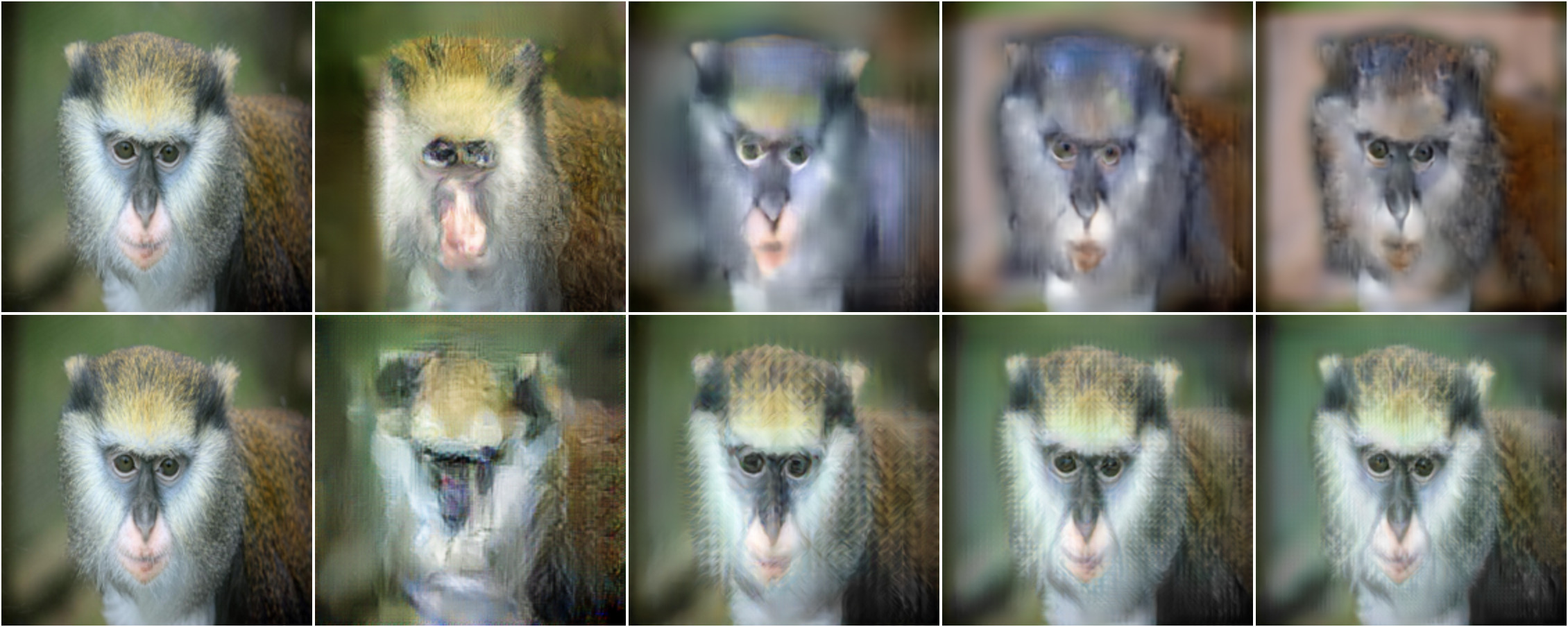}

\includegraphics[width=1\textwidth]{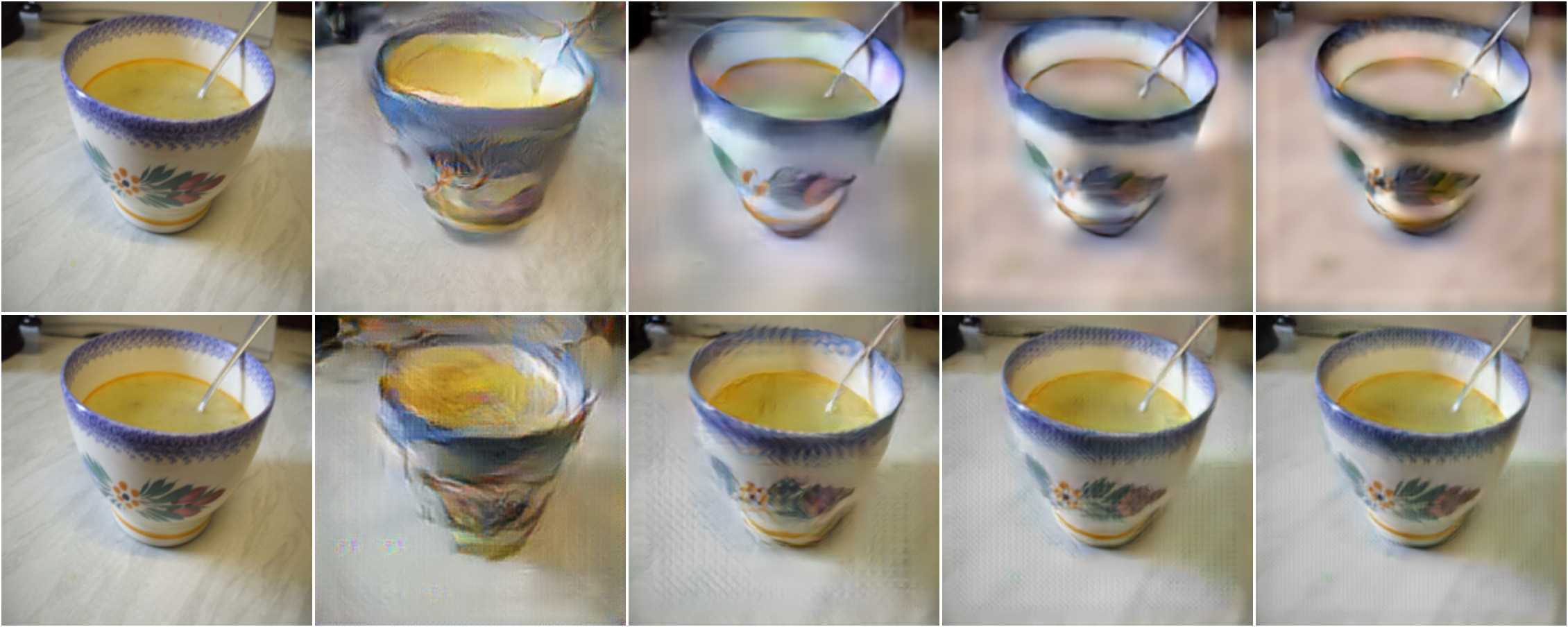}

\vspace{0.3 cm}
\captionof{figure}{\label{fig:eval_scale}Upscaled ImageNet samples reconstructed from their standard (top row) and AR (bottom row) features.}

\end{minipage}
\vspace{-0.5cm}
\end{table}

Results show LPIPS and SSIM improve almost monotonically until a maximum value is reached at $\varepsilon=2$, while PSNR keeps increasing. This implies that just by changing $\varepsilon$ from $0.5$ to $4$ while keeping the exact same architecture and training regime, a reconstruction improvement of $1.2$ dB PSNR is obtained.

Based on this, we use an AR AlexNet model trained with $\varepsilon=3$ in our experiments, which gives the best tradeoff between PSNR, SSIM and LPIPS. Overall, our analysis suggests that, while all four AR models outperform the inversion accuracy of the standard model, the reconstruction improvement is not proportional to the robustness level. Instead, it is maximized at a particular level. Please refer to \secref{sec:supp_inverting_alternative} for additional robustness level vs. reconstruction accuracy experiments on ResNet-18 pointing to the same conclusion.

\subsection{Reconstructing Images at Unseen Resolutions}
\label{sec:experimental_scale}

Unlike extracting shift-invariant representations, image scaling is difficult to handle for standard CNN-based models \cite{sosnovik_2019_scale,fan_2020_scale}. Following previous work suggesting AR features are more generic and transferable than standard ones \cite{chen2020shape,salman_2020_adversarially}, we test whether our proposed AR autoencoder generalizes better to scale changes. We explore this property and show that our model trained on low-resolution samples improves reconstruction of images at unseen scales without any fine-tuning.

\textbf{Scenario 1: Reconstructing Upscaled Images.} Upscaled ImageNet samples are reconstructed from their AR AlexNet \layer{conv5} representations. For a fair comparison across scales, each image is normalized to $224 \times 224$ px. and then enlarged by an integer factor $L>1$. Experiments show a higher accuracy obtained from AR features in terms of PSNR, SSIM and LPIPS (\tabref{tab:eval_scale}). All metrics improve almost monotonically with $L$. In contrast, accuracy using standard features degrades with $L$.
Inversion from AR features show almost perfect reconstruction for large scales, while those of standard features show severe distorsions (\figref{fig:eval_scale}).

\begin{table}[bpt!]
\begin{center}
\vspace{-0.2 cm}
\caption{\label{tab:hires} High-resolution images inverted using our AR AlexNet model (trained on low resolution images) show improved quality over standard inversions.}
\vspace{0.1 cm}
\fontsize{8.5}{10.5}\selectfont
\setlength\tabcolsep{2pt}
\resizebox{0.75\columnwidth}{!}{
\begin{tabular}{c|c|c|c} 
\specialrule{.15em}{.05em}{.05em} 
\makecell{Encoder} & PSNR (dB)$\uparrow$ & SSIM$\uparrow$ & LPIPS$\downarrow$\\
\hline
 Standard & \makecell{$14.266\pm$ $1.9015$} & \makecell{$0.3874\pm$ $0.151$} & \makecell{$0.5729\pm$ $0.0465$}\\
 AR (ours) & \makecell{$\mathbf{18.3606\pm}$ $\mathbf{2.6012}$} & \makecell{$\mathbf{0.4388\pm}$ $\mathbf{0.1508}$} & \makecell{$\mathbf{0.5673\pm}$ $\mathbf{0.0337}$}\\
\specialrule{.15em}{.05em}{.05em} 
\end{tabular}
}
\end{center}
\vspace{-0.625 cm}
\end{table}

\begin{figure}[t]
\begin{minipage}[t]{0.166\textwidth}
\centering\textbf{\colorbox{white}{\scalebox{.8}{Ground-truth}}}
\end{minipage}\begin{minipage}[t]{0.166\textwidth}
\centering \textbf{\colorbox{white}{\scalebox{.8}{Standard}}}
\end{minipage}\begin{minipage}[t]{0.1725\textwidth}
\centering \textbf{\colorbox{white}{\scalebox{.8}{AR (Ours)}}}
\end{minipage}\begin{minipage}[t]{0.166\textwidth}
\centering\textbf{\colorbox{white}{\scalebox{.8}{Ground-truth}}}
\end{minipage}\begin{minipage}[t]{0.166\textwidth}
\centering \textbf{\colorbox{white}{\scalebox{.8}{Standard}}}
\end{minipage}\begin{minipage}[t]{0.166\textwidth}
\centering \textbf{\colorbox{white}{\scalebox{.8}{AR (Ours)}}}
\end{minipage}

\vspace{-0.1 cm}    
\includegraphics[width=0.5\textwidth]{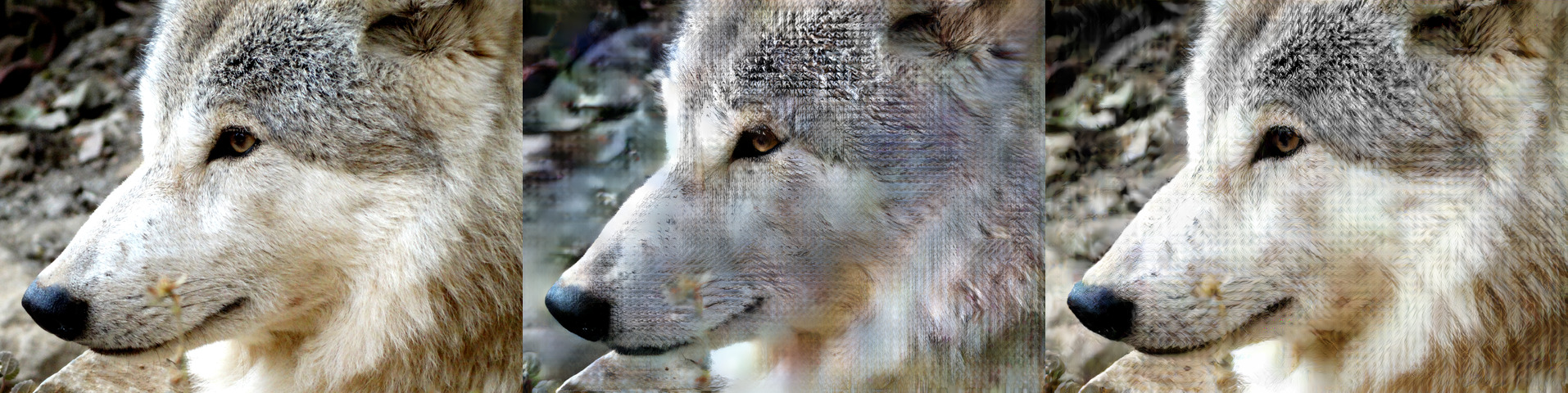}
\includegraphics[width=0.5\textwidth]{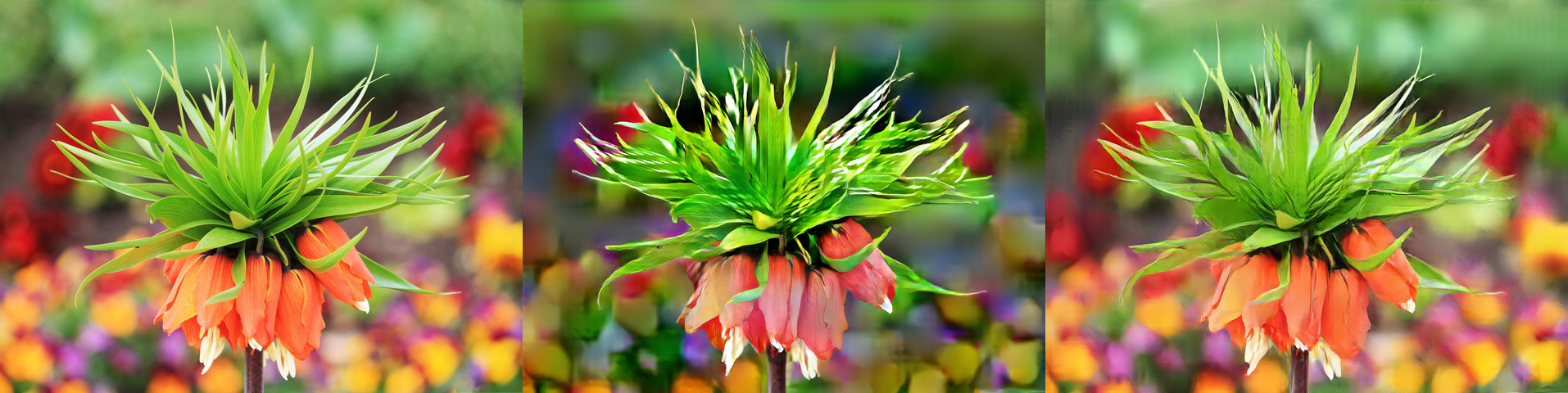}

\includegraphics[width=0.5\textwidth]{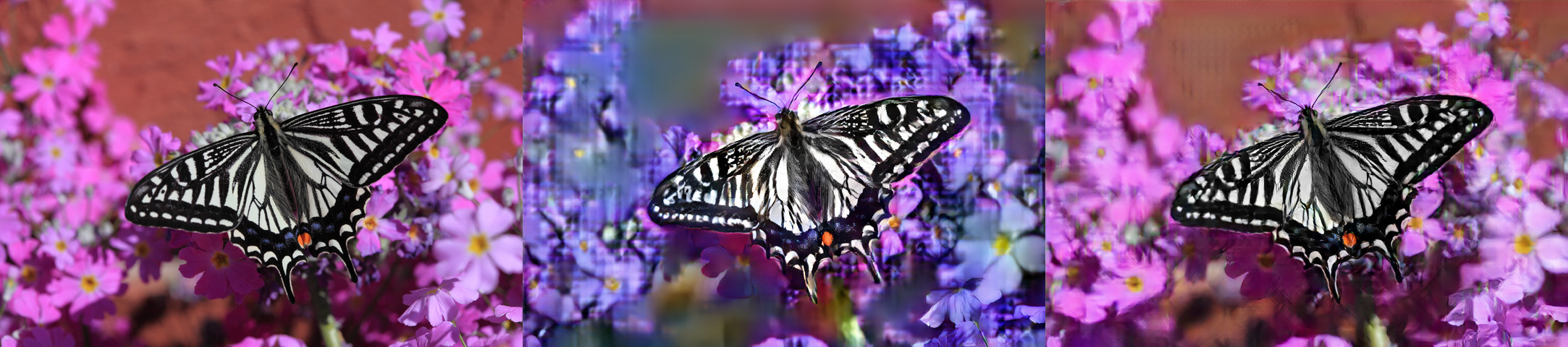}
\includegraphics[width=0.5\textwidth]{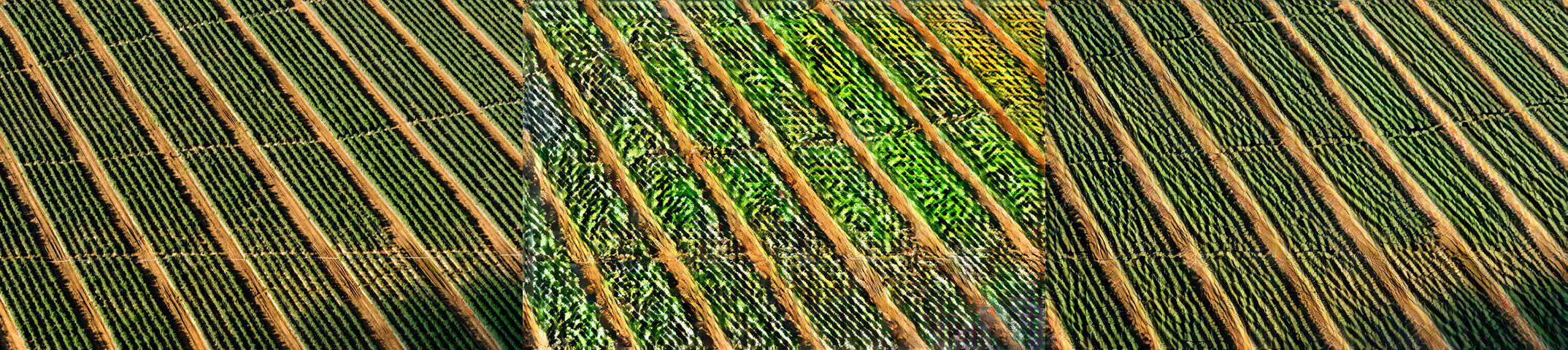}

\vspace{-0.1 cm}
\caption{\label{fig:hires}At a resolution of $2040\times 1536$ px., $10$ times larger than training samples, standard reconstructions on DIV2K show color and structure degradation. In contrast, reconstructions from our AR model do not suffer from distortions.}
\vspace{-0.5 cm}
\end{figure}

\textbf{Scenario 2: Reconstructing High-Resolution Images.} Standard and AR feature inversion is performed on the DIVerse 2K resolution dataset (DIV2K) \cite{agustsson_2017_ntire}, containing objects at multiple scales. AR feature reconstructions show a significant PSNR, SSIM and LPIPS improvement over standard ones, despite not being explicitly trained to handle such large-scale objects (\tabref{tab:hires}).

Qualitatively, reconstructions from AR AlexNet features preserve sharp edges, reduces color degradation and diminishes checkerboard effects induced by standard inversion (\figref{fig:hires}). Thus, for unseen scales and without finetuning, AR features better preserve structure without penalizing the perceptual similarity.

\subsection{Comparison against State-of-the-Art Inversion Techniques}
\label{sec:experimental_comparison}
The inversion accuracy of our AR autoencoder is compared against two alternative techniques: Optimization-based robust representation inversion (RI) \cite{engstrom_2019_adversarial} and DeePSiM \cite{dosovitskiy_2015_inverting}. For a fair comparison, all methods reconstruct images from AlexNet features. We begin by highlighting the differences between them.

While RI is a model-based approach that searches in the pixel domain for an image that matches a set of target AR features, we use a CNN-based generator trained on a combination of natural-image priors (\secref{sec:opt_crit}). On the other hand, while DeePSiM is also a CNN-based technique trained under multiple priors, its generator has approximately $63\%$ more trainable parameters than ours (\tabref{tab:inversion_comparison}).

\textbf{Experimental Setup.} All inversion methods are evaluated on ImageNet. Our standard and AR models are trained using pixel, feature and GAN losses using the training setup described in \secref{sec:supp_proposed_method}. DeePSiM is evaluated using its official Caffe implementation without any changes. RI is evaluated using its official PyTorch implementation modified to invert \layer{conv5} AR features. Input samples are rescaled to $224 \times 224$ px. ($227 \times 227$ px. for DeepSiM).

\textbf{Results.} Our AR AlexNet autoencoder obtains the best accuracy in terms of PSNR and the second best in terms of SSIM (\tabref{tab:inversion_comparison}). While it outperforms its standard version in PSNR and SSIM, it gets a marginally worse LPIPS. Moreover, our AR model outperforms RI in all metrics. Also, despite DeePSiM having more layers and using larger inputs, our model achieves a large PSNR improvement over it. Results highlight the improvement obtained by inverting AR features and how this fundamental change allows competitive reconstruction quality using three times less trainable parameters.

\begin{table}[t]
\centering
\vspace{-0.3cm}
\caption{\label{tab:inversion_comparison} Comparison against state-of-the-art inversion techniques. By inverting AR features, our autoencoder outperforms the optimization-based RI method by a large margin. Despite having 63\% less parameters, we also obtain favorable results against DeepSiM, showing a significant PSNR improvement.}%
\vspace{0.1 cm}
\resizebox{\columnwidth}{!}{

\begin{tabular}{c|c|c|c|c|c}
\specialrule{.15em}{.05em}{.05em} 
Algorithm & Encoder & Trainable Pars. & PSNR (dB)$\uparrow$ & SSIM$\uparrow$ & LPIPS$\downarrow$ \\
\hline
\makecell{RI \cite{engstrom_2019_adversarial}} & \makecell{AR AlexNet} & $-$ & $16.724\pm 2.434$ & $0.181\pm 0.071$ & $0.63\pm 0.04$\\
\makecell{Standard Autoencoder} & \makecell{Standard AlexNet} & $4,696,026$ & $15.057\pm 2.392$ & $0.307\pm 0.158$ & $\mathbf{0.547\pm 0.055}$\\
\makecell{AR Autoencoder (ours)} & \makecell{AR AlexNet} & $4,696,026$ & $\mathbf{17.227\pm 2.725}$ & $\mathbf{0.358\pm 0.163}$ & $0.567\pm 0.056$\\
\hline\hline
\makecell{DeepSiM \cite{dosovitskiy_2016_generating}} & \makecell{Standard CaffeNet} & $12,702,307$ & $15.321\pm 2.011$ & $0.417\pm 0.158$ & $0.531\pm 0.059$\\
\specialrule{.15em}{.05em}{.05em} 
\end{tabular}}
\vspace{-0.55 cm}
\end{table}

\section{Downstream Tasks}
\label{sec:tasks}

We further evaluate the benefits of incorporating AR autoencoders into two downstream tasks: style transfer and image denoising. To assess the benefits of AR autoencoders, in each task, we simply replace the standard autoencoders by the AR versions without incorporating any additional task-specific priors or tuning. Despite not tailoring our architecture to each scenario, it obtains on-par or better results than well-established methods. Refer to \secref{sec:supp_additional} and \secref{sec:supp_proposed_method} for more results and full implementation details.

\subsection{Style Transfer via Robust Feature Alignment}
\begin{table}[t]
\centering
\begin{center}
\vspace{0.2 cm}
\caption{\label{tab:style_quantitative}Universal Style Transfer. Our AR AlexNet autoencoder outperforms both its standard counterpart and the original VGG-19 model in terms of Gram loss, the latter using more layers, larger feature maps and feature blending.}
\vspace{0.1 cm}
\resizebox{0.75\textwidth}{!}{%
\begin{tabular}{c|c|c|c|c|c}
\specialrule{.15em}{.05em}{.05em} 
\makecell{Encoder} & \makecell{Stylization\\Levels} & \makecell{Smallest\\Feature Map} & \makecell{Feature\\Blending} & \makecell{Gram Loss$\downarrow$\\$(x_{cs}, x_{s})$} & \makecell{SSIM$\uparrow$\\$(x_{cs}, x_{c})$}\\
\hline
\makecell{Standard AlexNet} & $3$ & $6\times6\times256$ & \xmark & $1.694$ & $0.226$\\\hline
\makecell{AR AlexNet (ours)} & $3$ & $6\times6\times256$ & \xmark & $\mathbf{1.186}$ & $\mathbf{0.259}$\\
\hline\hline
\makecell{VGG-19 \cite{simonyan_2014_very}} & $5$ & $14\times14\times512$ & \cmark & ${1.223}$ & ${0.459}$\\
\specialrule{.15em}{.05em}{.05em} 
\end{tabular}}
\end{center}
\vspace{-0.5 cm}
\end{table}

\begin{figure}[t]
\begin{minipage}{0.1\textwidth}
\centering\textbf{\colorbox{white}{\scalebox{.8}{Refs}}}
\end{minipage}\begin{minipage}[t]{0.2\textwidth}
\centering \textbf{\colorbox{white}{\scalebox{.8}{Standard}}}
\end{minipage}\begin{minipage}[t]{0.21\textwidth}
\centering \textbf{\colorbox{white}{\scalebox{.8}{AR (ours)}}}
\end{minipage}\begin{minipage}[t]{0.1\textwidth}
\centering\textbf{\colorbox{white}{\scalebox{.8}{Refs}}}
\end{minipage}\begin{minipage}[t]{0.2\textwidth}
\centering \textbf{\colorbox{white}{\scalebox{.8}{Standard}}}
\end{minipage}\begin{minipage}[t]{0.2\textwidth}
\centering \textbf{\colorbox{white}{\scalebox{.8}{AR (ours)}}}
\end{minipage}

\vspace{-0.1 cm}
\includegraphics[width=0.5\textwidth]{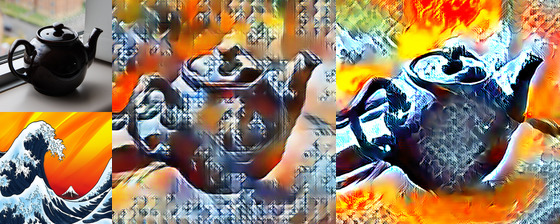}
\includegraphics[width=0.5\textwidth]{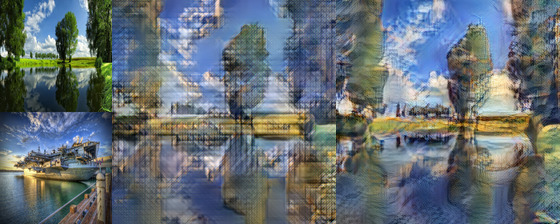}

\caption{\label{fig:style01}Universal Style Transfer: By inverting AR features, our autoencoder improves both content and style preservation, obtaining a better image stylization.}
\vspace{-0.7 cm}
\end{figure}

Motivated by the perceptual properties of AR features \cite{engstrom_2019_adversarial}, we analyze their impact on style transfer using our AR AlexNet autoencoder as backbone and measure their improvement in both structure and texture preservation.

\textbf{Experimental Setup.} Stylization is evaluated on $75$ random content images and $100$ random style images, leading to $7,500$ image pairs. Content and style preservation is evaluated via the SSIM between content and stylized images and the VGG-19 Gram loss between style and stylized images, respectively. \layer{Conv1} and \layer{conv2} models use nearest neighbor interpolation instead of transposed convolution layers to improve reconstruction and avoid checkerboard effects, while the \layer{conv5} model remains unaltered. We also include results using Universal Style Transfer's (UST) official implementation, using a VGG-19 backbone.

\textbf{Results.} Our AR autoencoder improves both texture and structure preservation over its standard version (\tabref{tab:style_quantitative}). Stylization via AR features removes artifacts in flat areas, reducing blurry outputs and degraded structure (\figref{fig:style01}). Besides, our AR model gets a lower Gram loss with respect to UST. This implies that, despite matching less feature maps than the VGG-19 model (three instead of five), stylizing via our AR AlexNet autoencoder better preserves the style.

As expected, UST obtains a better SSIM since VGG-19 has more complexity and uses less contracted feature maps than our AlexNet model (e.g. $14 \times 14 \times 512$ vs. $6 \times 6 \times 256$). Also, UST \textit{blends} stylized and content features to better preserve shapes. Overall, a comparison between our AR model and UST shows a tradeoff between content and style preservation.

\subsection{Image Denoising via AR Autoencoder}
Similarly to the robustness imposed by regularized autoencoders \cite{vincent_2010_stacked,rifai2011contractive,kingma_2013_auto}, we harness the manifold learned by AR models to obtain noise-free reconstructions. We evaluate our AR AlexNet denoising model and compare its restoration properties with alternative learn-based methods.
\begin{table}[t]
\vspace{0.3 cm}
\small
\def\arraystretch{1.25}
\setlength\tabcolsep{4pt}
\caption{\label{tab:denoising}Image denoising $(\sigma= \frac{50}{255})$: Our AR denoiser outperforms its standard version on multiple datasets. On the largest one (CBSD68), it also outperforms alternative learn-based techniques. On smaller sets (Kodak24, McMaster), it improves in SSIM and gets comparable PSNR and LPIPS performance.}
\vspace{-0.2 cm}
\begin{center}
\resizebox{\columnwidth}{!}{
\begin{tabular}{c|c|c|c|c|c|c|c|c|c}
\specialrule{.15em}{.05em}{.05em} 
& PSNR (dB)$\uparrow$ & SSIM$\uparrow$ & LPIPS$\downarrow$ & PSNR (dB)$\uparrow$ & SSIM$\uparrow$ & LPIPS$\downarrow$ & PSNR (dB)$\uparrow$ & SSIM$\uparrow$ & LPIPS$\downarrow$\\
\hline
Encoder & \multicolumn{3}{c|}{CBSD68} & \multicolumn{3}{c|}{Kodak24} & \multicolumn{3}{c}{McMaster}\\
\hline
\makecell{TNRD \cite{chen_2016_trainable}} & $24.75$ & $0.662$ & $0.445$ & $25.994$ & $0.695$ & $0.461$ & $25.01$ & $0.66$ & $\mathbf{0.387}$\\
\makecell{MLP \cite{burger_2012_image}} & $25.184$ & $0.663$ & $0.46$ & \makecell{$\mathbf{26.31}$} & $0.691$ & $0.478$ & \makecell{$\mathbf{26.039}$} & \makecell{$\mathbf{0.693}$} & \makecell{$0.402$}\\
\makecell{Standard} & \makecell{$22.6297$} & \makecell{$0.6178$} & \makecell{$0.567$} & \makecell{$23.1868$} & \makecell{$0.6001$} & \makecell{$0.4968$} & \makecell{$23.1493$} & \makecell{$0.6072$} & \makecell{$0.4458$}\\
\makecell{AR (ours)} & \makecell{$\mathbf{25.258}$} & \makecell{$\mathbf{0.7095}$} & \makecell{$\mathbf{0.4043}$} & \makecell{$25.4946$} & \makecell{$\mathbf{0.701}$} & \makecell{$\mathbf{0.447}$} & \makecell{$25.3527$}& \makecell{$0.6914$} & \makecell{$0.3965$}\\
\specialrule{.15em}{.05em}{.05em} 
\end{tabular}
}
\end{center}
\vspace{-0.7 cm}
\end{table}

\begin{figure}[t]
\begin{minipage}{0.25\textwidth}
\centering\textbf{\colorbox{white}{\scalebox{.8}{Ground-truth}}}
\end{minipage}\begin{minipage}[t]{0.25\textwidth}
\centering \textbf{\colorbox{white}{\scalebox{.8}{Observation}}}
\end{minipage}\begin{minipage}[t]{0.25\textwidth}
\centering \textbf{\colorbox{white}{\scalebox{.8}{Standard}}}
\end{minipage}\begin{minipage}[t]{0.25\textwidth}
\centering\textbf{\colorbox{white}{\scalebox{.8}{AR (ours)}}}
\end{minipage}

\vspace{-0.05 cm}
\includegraphics[width=1\textwidth]{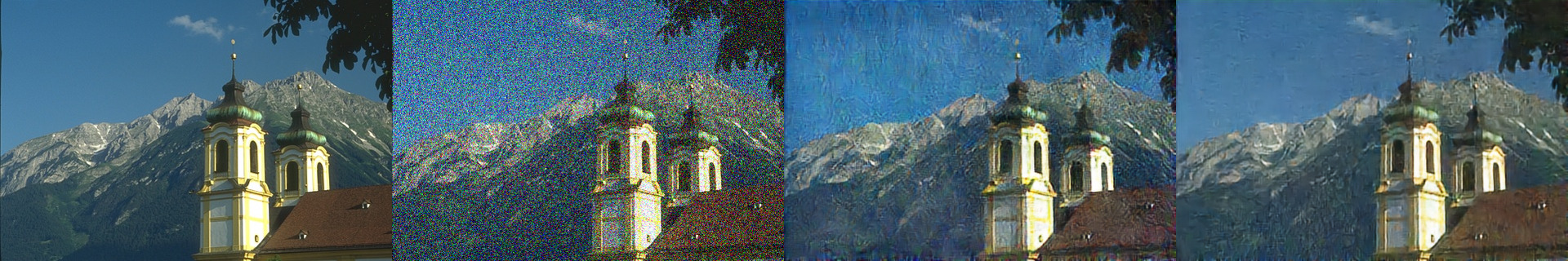}

\includegraphics[width=1\textwidth]{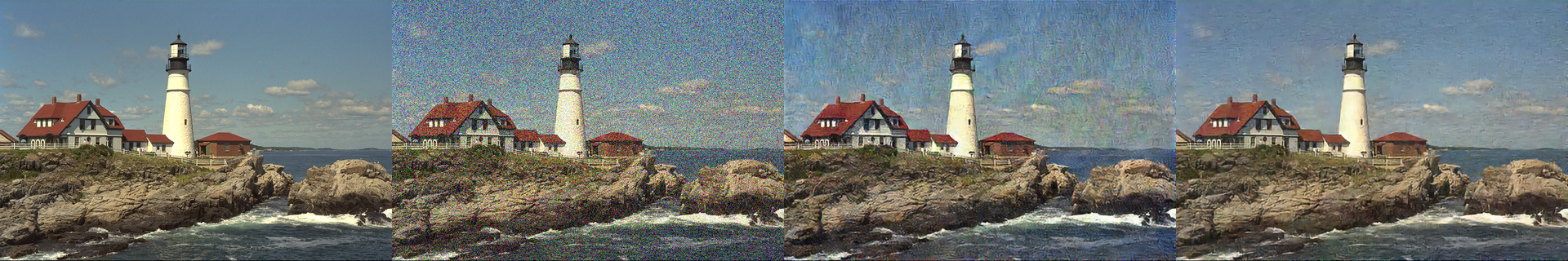}

\caption{\label{fig:denoising}Image denoising $(\sigma= \frac{50}{255})$: While inverting standard features introduces artifacts and degrades color, limiting their use for restoration tasks, our AR denoiser reduces the artifacts and better preserves the original texture.}
\vspace{-0.5 cm}
\end{figure}

\textbf{Experimental Setup.} Our image denoising model consists of an AR autoencoder equipped with skip connections in \layer{conv1}, \layer{conv2} and \layer{conv5} layers to better preserve image details. Skip connections follow the Wavelet Pooling approach \cite{yoo_2019_photorealistic}. Generators are trained on ImageNet via pixel and feature losses.

Accuracy is evaluated on the Kodak24, McMaster \cite{zhang_2011_color} and Color Berkeley Segmentation Dataset 68 (CBSD68) \cite{MartinFTM01} for clipped additive Gaussian noise ($\sigma=50/255$). We compare our AR model against two learn-based methods, Trainable Nonlinear Reaction Diffusion (TNRD) \cite{chen_2016_trainable} and Multi Layer Perceptron-based model (MLP) \cite{burger_2012_image}, often included in real-noise denoising benchmarks \cite{anwar_2019_real,guo_2019_toward}.

\textbf{Results.}  Our AR model improves over its standard version in all metrics across all datasets (\tabref{tab:denoising}). While standard predictions include color distorsions and texture artifacts, AR predictions show a better texture preservation and significantly reduce the distorsions introduced by the denoising process (\figref{fig:denoising}).

Our AR model obtains the best PSNR, SSIM and LPIPS scores on CBSD68, the most diverse of all datasets. While it is outperformed in PSNR by MLP in the two remaining datasets, it improves in SSIM and LPIPS, getting best or second best performance. For the McMaster dataset, SSIM and LPIPS values obtained by our model are slightly below the best values. Overall, our model consistently preserves the perceptual and structural similarity across all datasets, showing competitive results with alternative data-driven approaches.

\section{Conclusions}
\label{sec:conclusions}
A novel encoding-decoding model for synthesis tasks is proposed by exploiting the perceptual properties of AR features. We show the reconstruction improvement obtained by generators trained on AR features and how it generalizes to models of different complexity. We showcase our model on style transfer and image denoising tasks, outperforming standard approaches and attaining competitive performance against alternative methods. A potential limitation of our model is the loss of details due to its contracted features. Yet, experiments show that using shortcut connections allow preserving these, enabling enhancement and restoration tasks. Our method also requires pre-training an AR encoder prior to training the generator, which may increase its computational requirements.

Learning how to invert AR features may be interestingly extended to conditional GANs for image-to-image translation tasks \cite{isola2017image} and to VAEs as a latent variable regularizer \cite{dosovitskiy_2016_generating}. Our AR autoencoder can also be seen as an energy-based model \cite{nguyen2017plug} for artificial and biological neural networks vizualization \cite{nguyen2016synthesizing,nguyen2019understanding,ponce2019evolving}.

\vspace{\baselineskip}
{\noindent \textbf{Acknowledgements.}}
AN was supported by NSF Grant No. 1850117 \& 2145767, and donations from NaphCare Foundation \& Adobe Research.
We are grateful for Kelly Price's tireless assistance with our GPU servers at Auburn University.

\bibliographystyle{splncs}
\bibliography{refs}

\clearpage
\newcommand{\beginsupplementary}{
    \setcounter{section}{0}
	\renewcommand{\thesection}{A\arabic{section}}
	\renewcommand{\thesubsection}{\thesection.\arabic{subsection}}

	\renewcommand{\thetable}{A\arabic{table}}
	\setcounter{table}{0}

	\renewcommand{\thefigure}{A\arabic{figure}}
	\setcounter{figure}{0}
}

\newcommand{\suptitle}{Appendix for:\\ \papertitle}

\newcommand{\maketitlesupp}{
    \vskip .375in
    \begin{center}
        \Large \bf \suptitle \par
    \end{center}
    \vspace{24pt}}
\beginsupplementary
\maketitlesupp

{\noindent The appendix is organized as follows:}
\begin{itemize}
\item In~\secref{sec:supp_anomaly_detection}, we present \textbf{a third application}, GAN-based \textit{One-vs-All} anomaly detection using AR features, and show its benefits over standard techniques.
\item In~\secref{sec:supp_results}, we provide additional experimental results on feature inversion.
\item In~\secref{sec:supp_additional}, we provide additional experimental results on downstream tasks. 
\item In~\secref{sec:supp_proposed_method}, we provide implementation and experimental setup details.
\end{itemize}

\section{Anomaly Detection using AR Representations}
\label{sec:supp_anomaly_detection}
\subsection{Approach}
\textit{One-vs-All} anomaly detection is the task of identifying samples that do not fit an expected pattern \cite{golan_2018_deep,deecke_2018_image,ruff_2019_deep,wang_2019_effective}. Given an unlabeled image dataset with normal (\textit{positives}) and anomalous instances (\textit{negatives}), the goal is to distinguish between them. Following GAN-based techniques \cite{deecke_2018_image}, we train our proposed AR AlexNet autoencoder exclusively on positives to learn a how to accurately reconstruct them. Once trained on such a target distribution, we use its reconstruction accuracy to detect negatives.

\begin{figure}[t]
\vspace{-0.2cm}
\subfloat[Reconstruction-based Anomaly Score]{\includegraphics[width=0.475\textwidth]{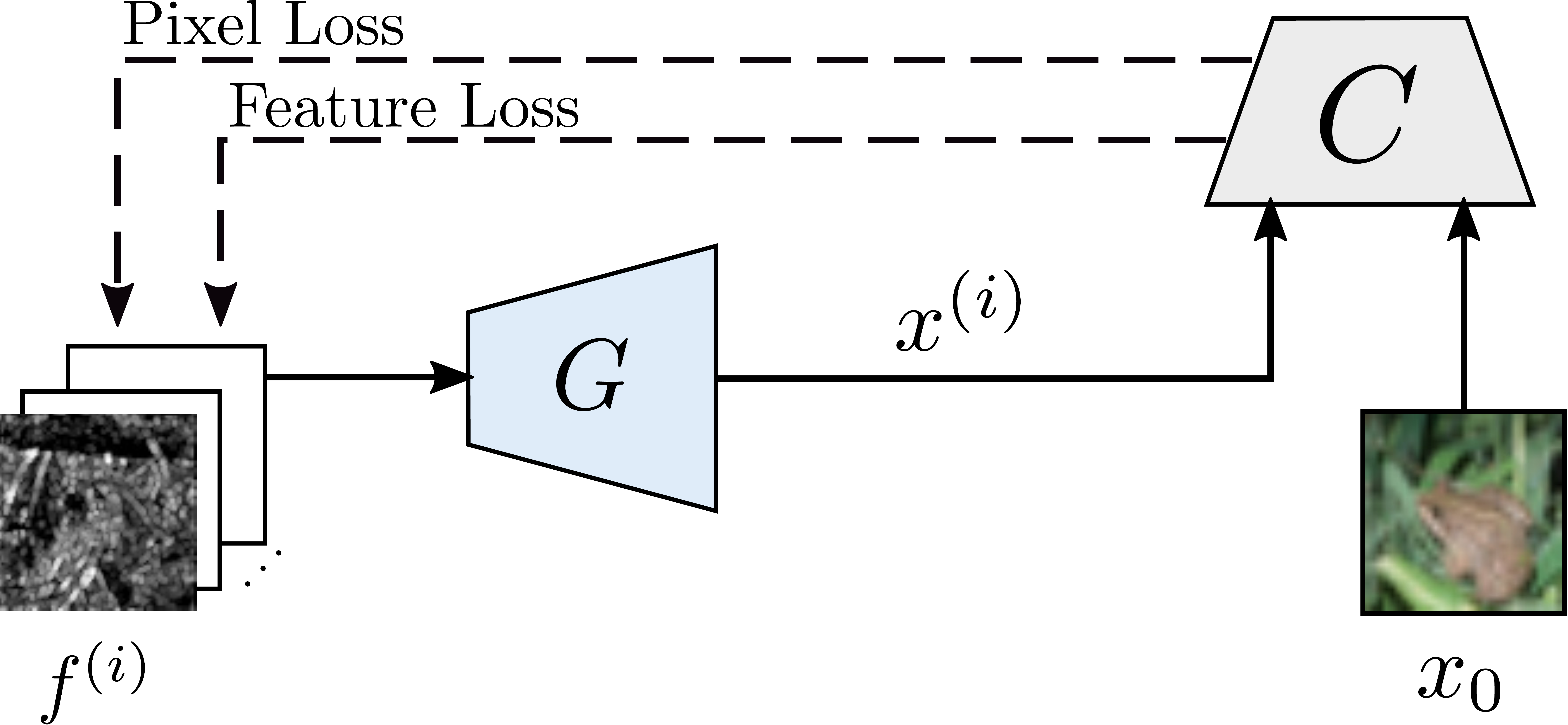}}\hspace{0.05\textwidth}
\subfloat[\textit{One-vs-All} Anomaly Detection]{\includegraphics[width=0.405\textwidth]{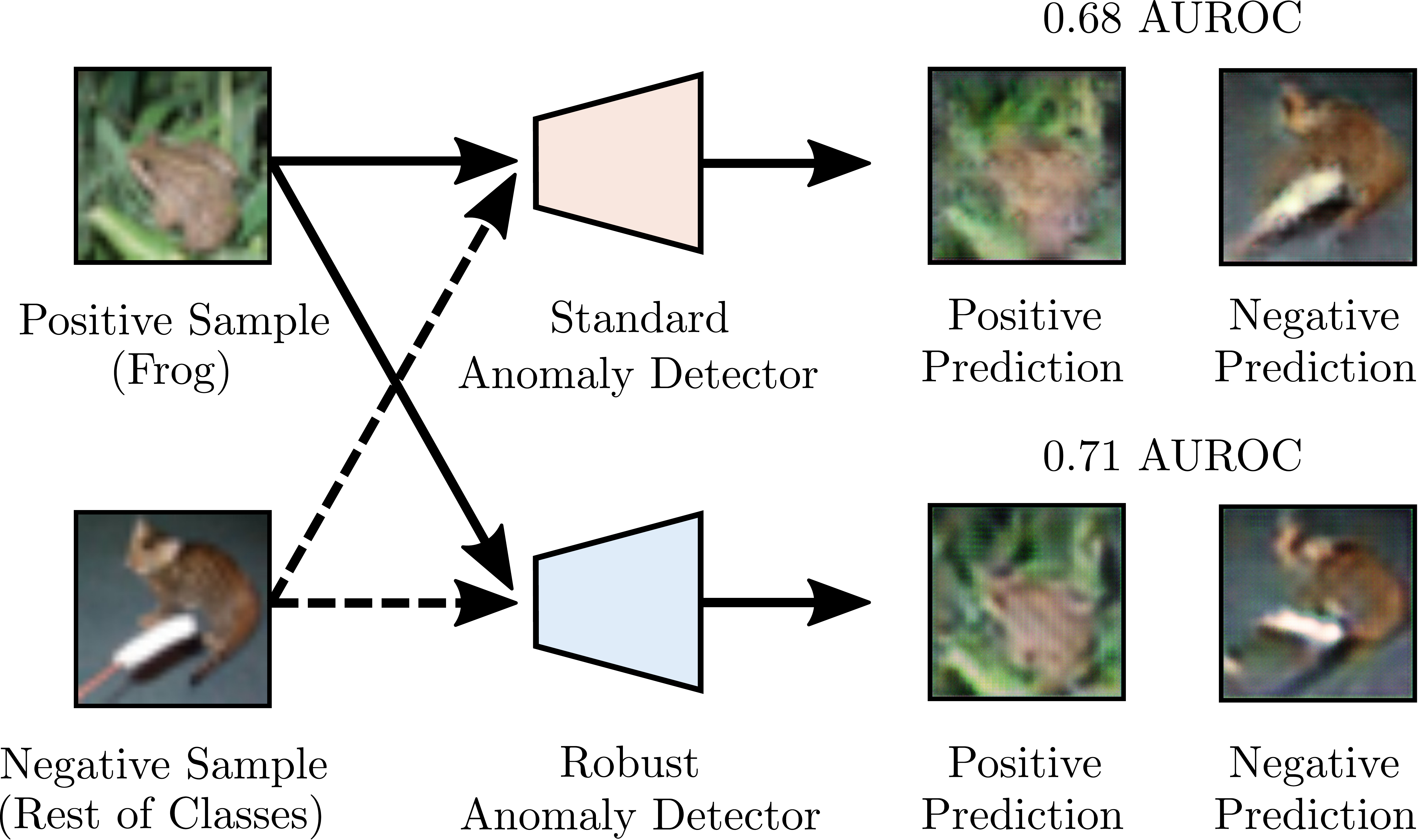}}
\vspace{-0.2cm}
\caption{\label{fig:proposed_anomaly}Anomaly Detection using adversarially robust features.}
\vspace{-0.4 cm}
\end{figure}

Given an unlabeled sample $x$ and its AR features $f$, we search for $\hat{f}$ that yields the best reconstruction $\hat{x}=G_{\tilde{\phi}}(\hat{f})$ based on the following criterion (\figref{fig:proposed_anomaly}):
\begin{align}
    \label{eq:anomaly_detection}
    \hat{f}=\ \text{arg }\underset{f}{\text{min}}&\ \alpha_{\text{pix}}\|G_{\tilde{\phi}}(f)-x\|_{1}+ \alpha_{\text{feat}}\|F_{\tilde{\theta}}\circ G_{\tilde{\phi}}(f)-F_{\tilde{\theta}}(x)\|_{2}^{2},
\end{align}
where $\alpha_{\text{pix}}, \alpha_{\text{feat}} \in \mathbb{R}_{++}$ are hyperparameters. Essentially, $x$ is associated to $\hat{f}$ that minimizes pixel and feature losses between estimated and target representations. Since $G_{\tilde{\phi}}$ has been trained on the distribution of positive samples, latent codes of negative samples generate abnormal reconstructions, revealing anomalous instances.

\subsection{Experiments}
We hypothesize that our AR generator widens the reconstruction gap between in and out-of-distribution samples, improving its performance on anomaly detection. Given a labeled dataset, our generator is trained to invert AR features from samples of a single class (\textit{positives}). Then, we evaluate how accurately samples from the rest of classes (\textit{negatives}) are distinguished from positives on an unlabeled test set.

\textbf{Experimental Setup.} We compare our technique using AR and standard features against ADGAN \cite{deecke_2018_image,golan_2018_deep}. We evaluate the performance on CIFAR10 and Cats vs. Dogs \cite{parkhi_2012_cats} datasets, where AUROC is computed on their full test sets.

Standard and AR encoders are fully-trained on ImageNet using the parameters described in \secref{sec:supp_results}. By freezing the encoder, generators are trained using pixel and feature losses on positives from the dataset of interest, CIFAR10 or Cats vs. Dogs. Input images are rescaled to $224 \times 224$ px. before being passed to the model, no additional data augmentation is applied during the generator training. The regularization parameters for both standard and AR autoencoders are heuristically selected as:
\vspace{-0.1\baselineskip}
\begin{itemize}
    \setlength\itemsep{0.1\baselineskip}
    \item Standard autoencoder: $\lambda_{\text{pix}}=2\times 10^{-3},\; \lambda_{\text{feat}}=1\times 10^{-2}$.
    \item AR autoencoder: $\lambda_{\text{pix}}=2\times 10^{-6},\; \lambda_{\text{feat}}=1\times 10^{-2}$.
\end{itemize}

\textbf{Iterative Optimization Details.} After training the generator on a particular class of interest, the optimal latent code $\hat{f}$ associated to an arbitrary target image $x$ is obtained via stochastic gradient descent. For both standard and AR autoencoders, the optimization criteria are identical to that used during the generator training. Specifically, we minimize pixel and feature loss components using the following hyperparameters:
\vspace{-0.1\baselineskip}
\begin{itemize}
    \setlength\itemsep{0.1\baselineskip}
    \item Standard autoencoder: $\alpha_{\text{pix}}=2\times 10^{-3},\; \alpha_{\text{feat}}=1\times 10^{-2}$.
    \item AR autoencoder: $\alpha_{\text{pix}}=2\times 10^{-6},\; \alpha_{\text{feat}}=1\times 10^{-2}$.
\end{itemize}

Detection is performed by solving Eq. \eqref{eq:anomaly_detection}, where $f\in \mathbb{R}^{6 \times 6 \times 256}$ is initialized as white Gaussian noise and optimized for $i_{\text{max}}=100$ iterations. The initial learn rate is chosen as $0.1$ and linearly decreases along iterations down to $0.001$.

\textbf{Results.} Full \textit{one-vs-all} anomaly detection results for CIFAR-10 and Cats vs. Dogs datasets are shown in \tabref{tab:supp_anomaly_detection}. On average, our AR model improves on outlier detection over its standard version and ADGAN. Our AR model gets $6.51\%$ and $8.84\%$ relative AUROC improvement over ADGAN on CIFAR-10 and Cats vs. Dogs, respectively. This shows our generator better distinguishes positives and negatives due to its improved reconstruction accuracy.
\begin{table}[t]
\small
\begin{center}
\resizebox{0.65\textwidth}{!}{
\begin{tabular}{c|c|c|c|c} 
\specialrule{.15em}{.05em}{.05em} 
Dataset & \makecell{Positive\\Class} & \makecell{ADGAN\\ \cite{deecke_2018_image}} & \makecell{Proposed\\(Standard)} & \makecell{Proposed\\(AR)}\\
\hline
\multirow{11}{*}{CIFAR-10} & $0$ & $0.649$ & $0.6874$ & $0.6533$\\
                        & $1$ & $0.39$ & $0.3498$ & $0.3755$\\
                        & $2$ & $0.652$ & $0.6756$ & $0.662$\\
                        & $3$ & $0.481$ & $0.5708$ & $0.6123$\\
                        & $4$ & $0.735$ & $0.751$ & $0.7538$\\
                        & $5$ & $0.476$ & $0.5101$ & $0.5278$\\
                        & $6$ & $0.623$ & $0.6895$ & $0.7113$\\
                        & $7$ & $0.487$ & $0.4773$ & $0.4526$\\
                        & $8$ & $0.66$ & $0.7232$ & $0.7008$\\
                        & $9$ & $0.378$ & $0.362$ & $0.4408$\\
\cline{2-5}
                        & Average & $0.553$ & $0.5797$ & $\mathbf{0.589}$\\
\hline
\multirow{3}{*}{\makecell{Cats vs. Dogs}}  & $0$ & $0.507$ & $0.663$ & $0.649$\\
                                            & $1$ & $0.481$ & $0.392$ & $0.427$\\
\cline{2-5}
                                            & Average & $0.494$ & $0.527$ & $\mathbf{0.538}$\\
\specialrule{.15em}{.05em}{.05em} 
\end{tabular}}
\end{center}
\vspace{-0.1 cm}
\caption{\label{tab:supp_anomaly_detection} AUROC of our proposed \textit{one-versus-all} anomaly detection method for each class. Detection evaluated on CIFAR-10 and Cats vs. Dogs datasets. Best results highlighted in black.}
\vspace{-0.5 cm}
\end{table}

\section{Additional Experiments on Feature Inversion}
\label{sec:supp_results}

\subsection{Ablation Study}
\label{sec:supp_results_ablation}

\begin{figure*}[t]
\subfloat[Ground-truth images.]{\includegraphics[width=\textwidth]{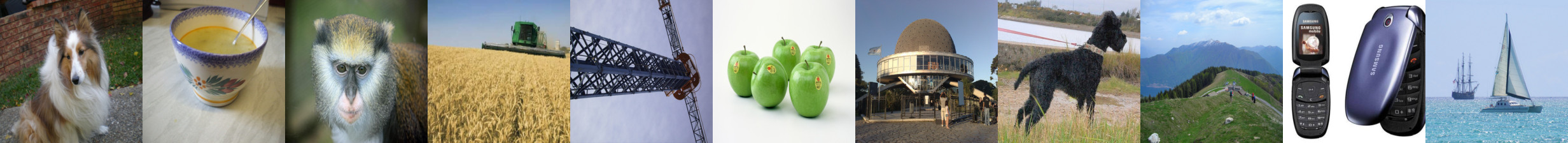}}

\vspace{-0.9\baselineskip}
\subfloat[Inverting standard (top) and AR (bottom) features using pixel losses.]{\includegraphics[width=\textwidth]{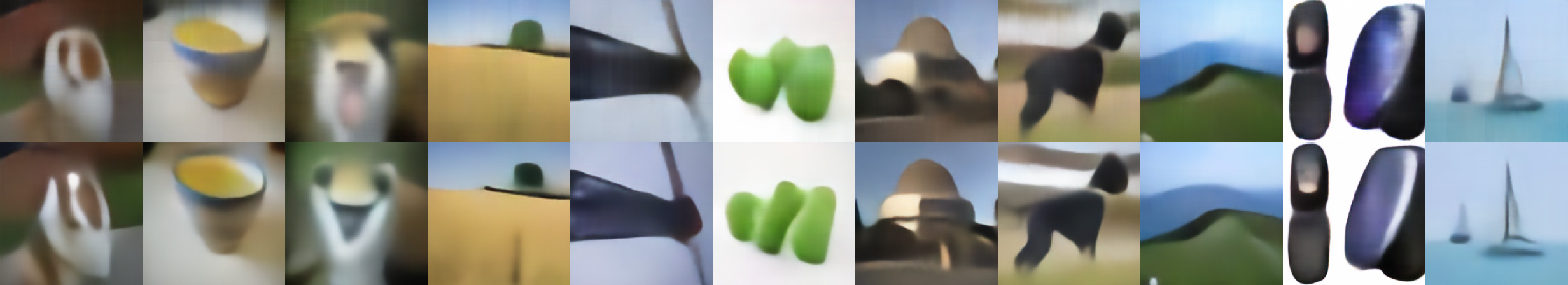}}

\vspace{-0.9\baselineskip}
\subfloat[Inverting standard (top) and AR (bottom) features using pixel and feature losses.]{\includegraphics[width=\textwidth]{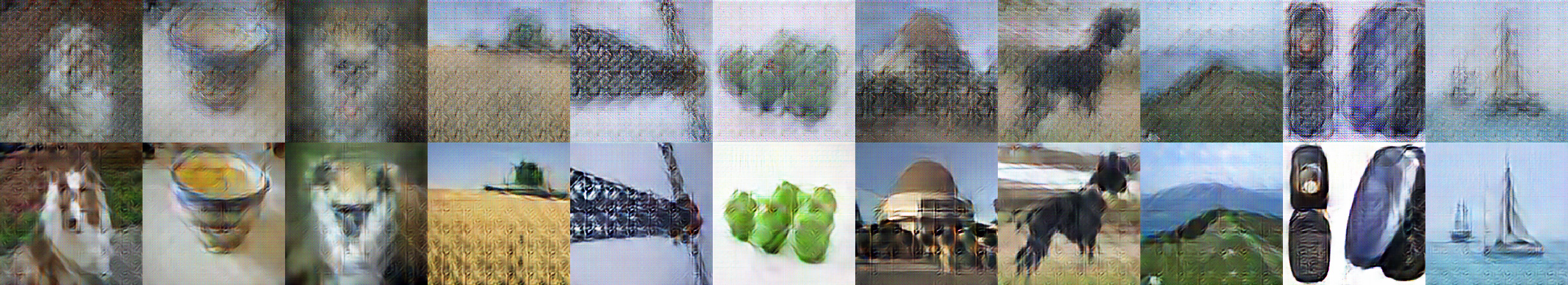}}

\vspace{-0.9\baselineskip}
\subfloat[Inverting standard (top) and AR (bottom) features using pixel, feature and GAN losses.]{\includegraphics[width=\textwidth]{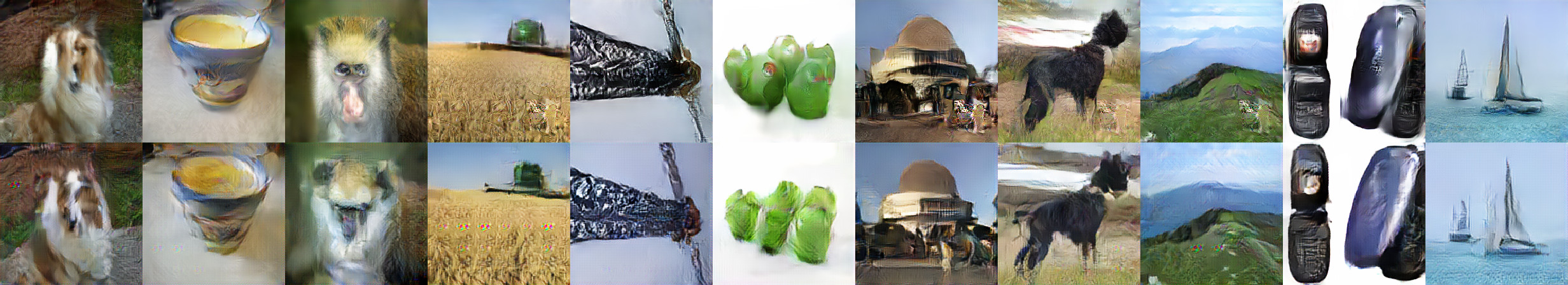}}

\caption{CNN-based feature inversion of standard and AR representations. AlexNet \layer{Conv5} \textbf{standard (top)} and \textbf{AR (bottom)} features are inverted using an image generator trained on (a) $\ell_{1}$ Pixel loss, (b) Pixel and feature losses, and (c) Pixel, feature and GAN losses.}
\label{fig:supp_ablation}
\end{figure*}

Feature inversion results obtained using different optimization criteria are illustrated in \figref{fig:supp_ablation}. Results clearly show the effect of each term, $\ell_{1}$ pixel, feature and GAN components, in the final reconstruction. Samples correspond to the ImageNet validation set. Particularly, when inverting features using pixel and feature losses, adversarially robust features show a significant improvement with respect to their standard counterparts. This agrees with the idea of adversarially robust features being perceptually aligned.

\subsection{Robustness to Scale Changes}
\label{sec:supp_results_scale}
\begin{figure*}[t]
\centering
\begin{minipage}{0.09\textwidth}
\centering\textbf{\hspace{-0.5\baselineskip}\scalebox{0.7}{G. truth}}
\end{minipage}\begin{minipage}{0.09\textwidth}
\centering \scalebox{0.7}{$L= 1$}
\end{minipage}\begin{minipage}{0.09\textwidth}
\centering \scalebox{0.7}{$L= 2$}
\end{minipage}\begin{minipage}{0.09\textwidth}
\centering \scalebox{0.7}{$L= 3$}
\end{minipage}\begin{minipage}{0.09\textwidth}
\centering \scalebox{0.7}{$L= 4$}
\end{minipage}\begin{minipage}{0.09\textwidth}
\centering \scalebox{0.7}{$L= 5$}
\end{minipage}\begin{minipage}{0.09\textwidth}
\centering \scalebox{0.7}{$L= 6$}
\end{minipage}\begin{minipage}{0.09\textwidth}
\centering \scalebox{0.7}{$L= 7$}
\end{minipage}\begin{minipage}{0.09\textwidth}
\centering \scalebox{0.7}{$L= 8$}
\end{minipage}\begin{minipage}{0.09\textwidth}
\centering \scalebox{0.7}{$L= 9$}
\end{minipage}\begin{minipage}{0.09\textwidth}
\centering \scalebox{0.7}{$L= 10$}
\end{minipage}\vspace{0.5\baselineskip}

\vspace{-0.5 cm}
\subfloat{\includegraphics[width=1\textwidth]{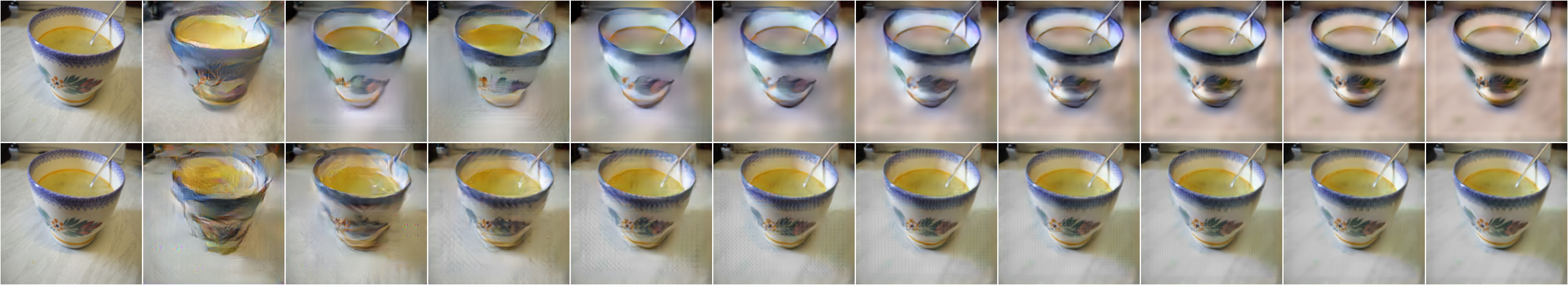}}

\vspace{-.9\baselineskip}
\subfloat{\includegraphics[width=1\textwidth]{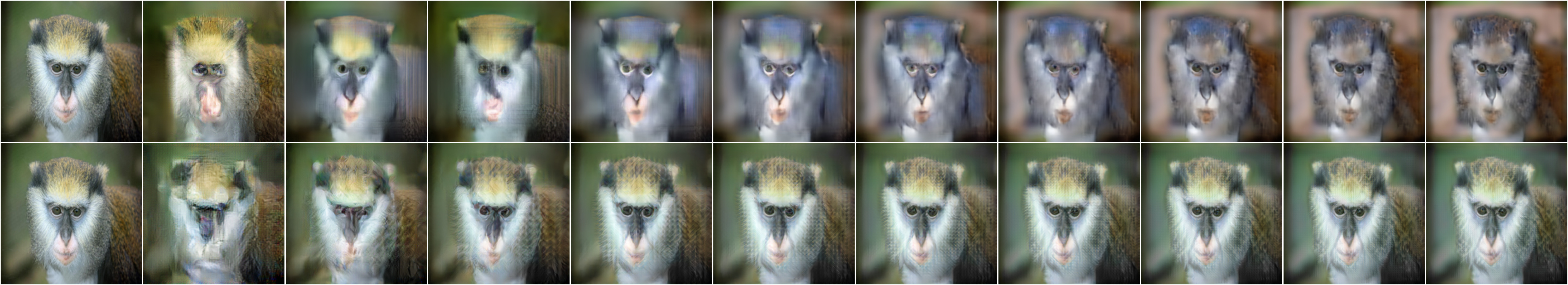}}

\vspace{-.9\baselineskip}
\subfloat{\includegraphics[width=1\textwidth]{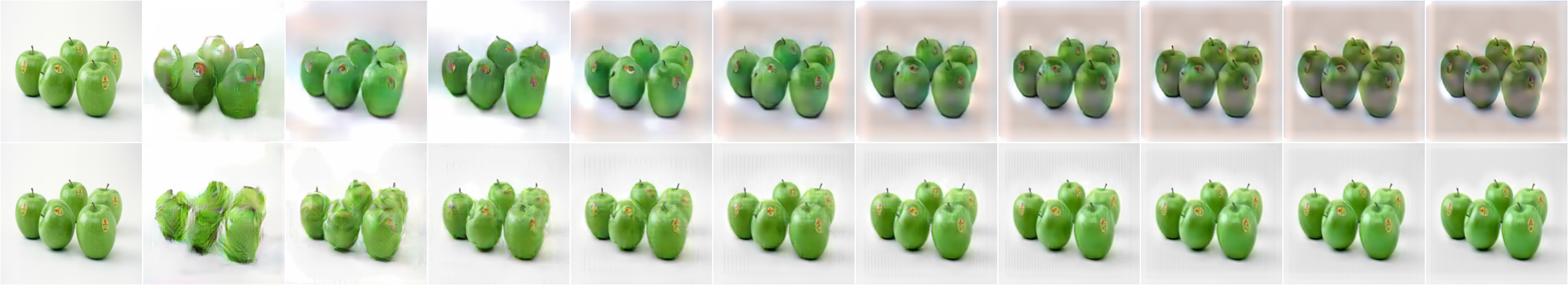}}

\vspace{-.9\baselineskip}
\subfloat{\includegraphics[width=1\textwidth]{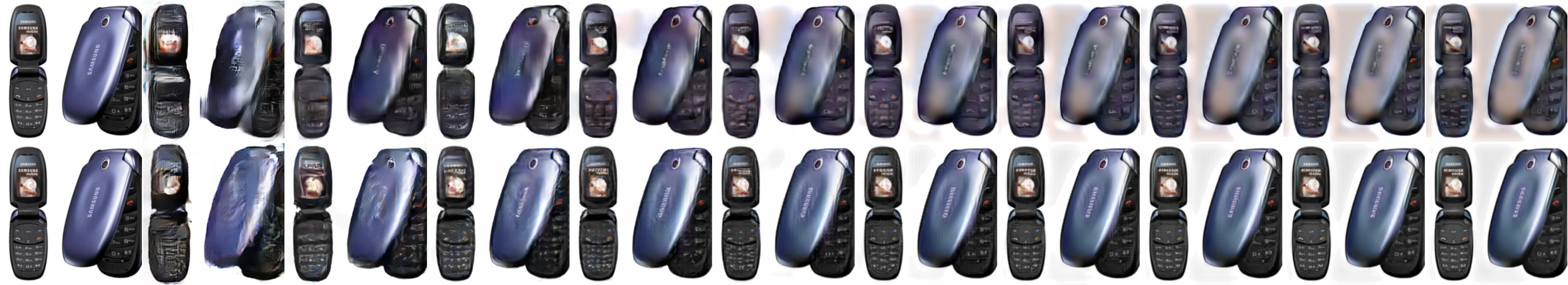}}

\caption{Reconstructing upscaled images. Upscaled ImageNet samples are inverted from their standard and AR representations. While standard representations (top row) are severely degraded, AR representations (bottom row) show an outstanding accuracy that improves with the scaling factor.}
\label{fig:sup_hires01}
\end{figure*}

Inversion accuracy on upscaled low-resolution images is illustrated in \figref{fig:sup_hires01} for scale factors $L\in \{1,\dots, 10\}$. While standard inversions show significant distortions for large upscaling factors $L$, reconstructions from adversarially robust representations show almost perfect reconstruction for high upscaling factors. Quantitative results are included in \tabref{tab:supp_hires01}. Results improve almost monotonically when inverting AR representations, even without exposing the Autoencoder to high-resolution images during training and without any fine-tuning.

\begin{figure*}[t]
\begin{minipage}[t]{0.166\textwidth}
\centering\textbf{\scriptsize{Ground-truth}}
\end{minipage}\begin{minipage}[t]{0.166\textwidth}
\centering \textbf{\scriptsize{Standard}}
\end{minipage}\begin{minipage}[t]{0.166\textwidth}
\centering \textbf{\scriptsize{AR (ours)}}
\end{minipage}\begin{minipage}[t]{0.166\textwidth}
\centering\textbf{\scriptsize{Ground-truth}}
\end{minipage}\begin{minipage}[t]{0.166\textwidth}
\centering \textbf{\scriptsize{Standard}}
\end{minipage}\begin{minipage}[t]{0.166\textwidth}
\centering \textbf{\scriptsize{AR (ours)}}
\end{minipage}

\vspace{-0.8\baselineskip}
\subfloat[]{\includegraphics[width=0.5\textwidth]{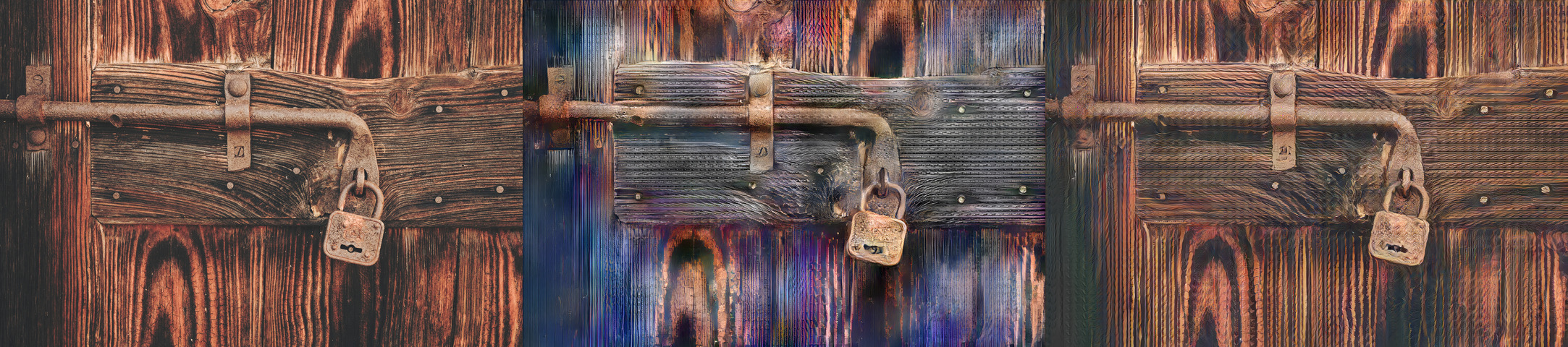}}
\subfloat[]{\includegraphics[width=0.5\textwidth]{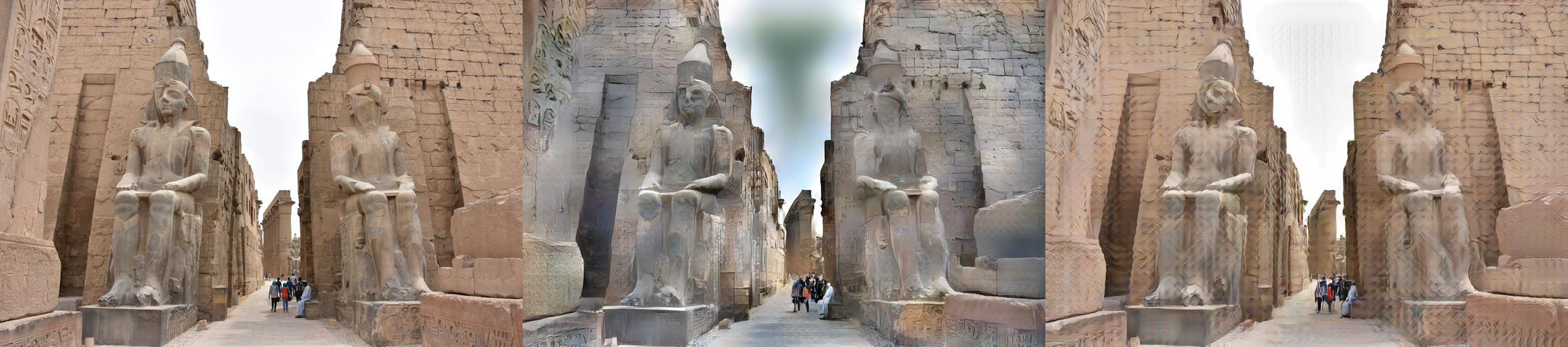}}

\vspace{-.9\baselineskip}
\subfloat[]{\includegraphics[width=0.472\textwidth]{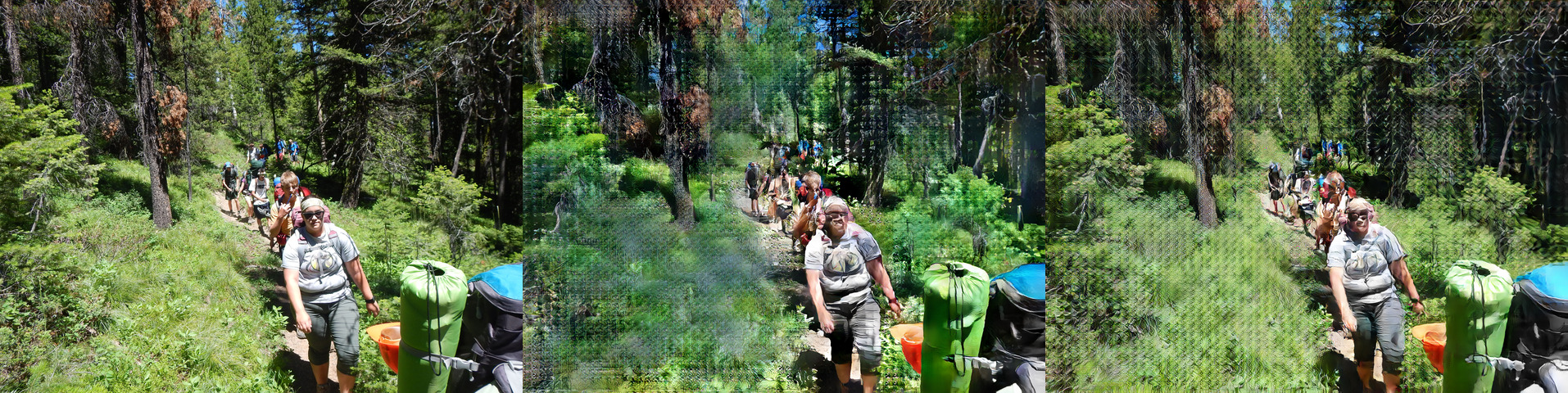}}
\subfloat[]{\includegraphics[width=0.528\textwidth]{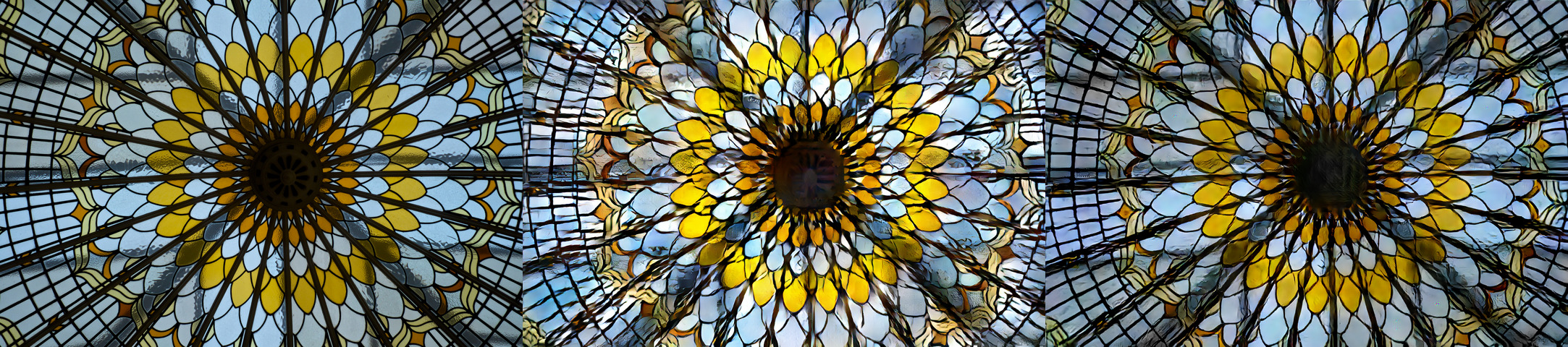}}

\vspace{-.9\baselineskip}
\subfloat[]{\includegraphics[width=0.45\textwidth]{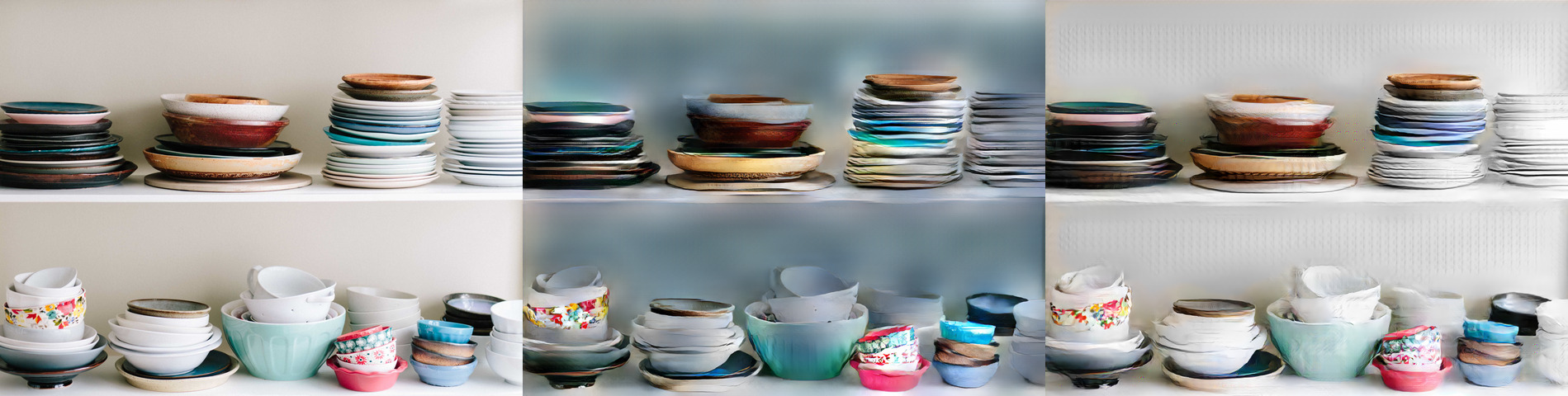}}
\subfloat[]{\includegraphics[width=0.55\textwidth]{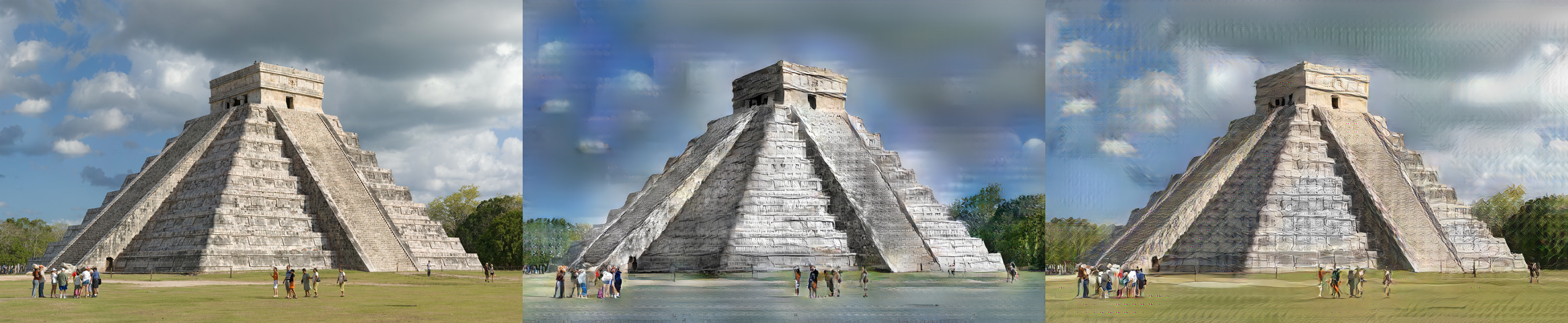}}

\vspace{-.9\baselineskip}
\subfloat[]{\includegraphics[width=0.5\textwidth]{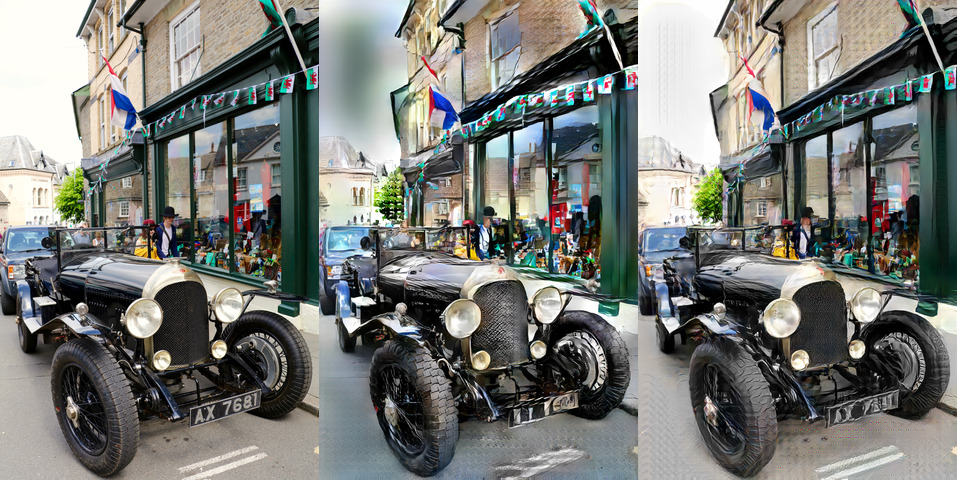}}
\subfloat[]{\includegraphics[width=0.5\textwidth]{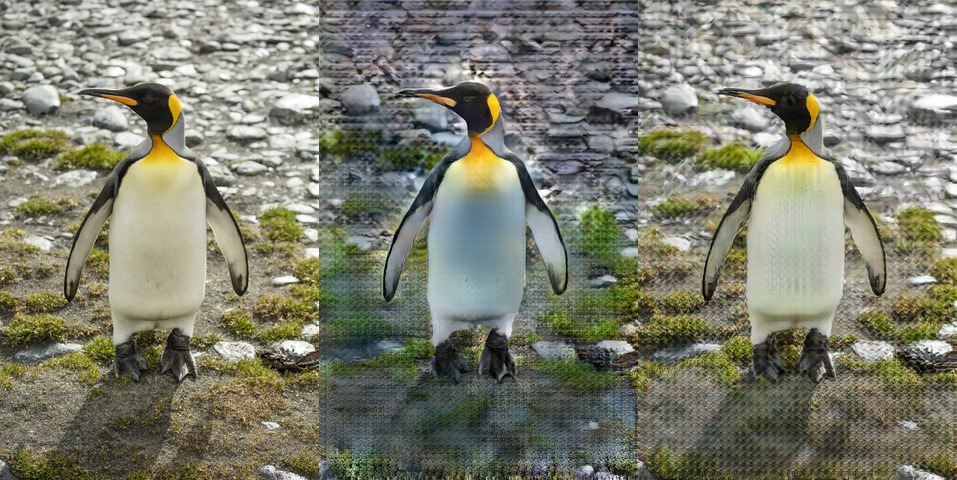}}

\vspace{-.3 cm}
\caption{At a resolution of $2040\times 1536$, $10$ times higher than the training resolution, standard reconstructions show color and structure degradation. In contrast, reconstructions from our AR autoencoder do not suffer from such distortions and are closer to target DIV2K images.}
\label{fig:supp_hires02}
\end{figure*}

On the other hand, extended results on feature inversion from high-resolution images are illustrated in \figref{fig:supp_hires02}. Notice that, in contrast to the previous case, input samples correspond to natural high-resolution images and are encoded without any scaling. Results show a good color and edge preservation from our AR autoencoder, while inverting standard features show bogus components and noticeable color distortions.

\begin{table*}[t]
\small
\centering
\begin{center}
\resizebox{0.9\textwidth}{!}{
\def\arraystretch{1.25}
\begin{tabular}{c|c|c|c|c|c|c}
\specialrule{.15em}{.05em}{.05em} 
\multirow{2}{*}{\makecell{$L$}} & \multicolumn{3}{c|}{Standard AlexNet} & \multicolumn{3}{c}{Robust AlexNet} \\
\cline{2-7}
& PSNR (dB)$\uparrow$ & SSIM$\uparrow$ & LPIPS$\downarrow$ & PSNR (dB)$\uparrow$ & SSIM$\uparrow$ & LPIPS$\downarrow$\\
\hline
 $1\ (224 \times 224)$ & \makecell{$15.057$} & \makecell{$0.3067$} & \makecell{$0.5473$} & \makecell{$17.2273$} & \makecell{$0.3580$} & \makecell{$0.5665$}\\
 $2\ (448 \times 448)$ & $16.2777$ & $0.4068$ & $0.4234$ & $20.3554$ & $0.4859$ & $0.469$\\
 $3\ (672 \times 672)$ & $16.0668$ & $0.4317$ & $0.4143$ & $21.3696$ & $0.5265$ & $0.4376$\\
 $4\ (896 \times 896)$ & \makecell{$15.4258$} & \makecell{$0.4655$} & \makecell{$0.4136$} & \makecell{$22.575$} & \makecell{$0.5892$} & \makecell{$0.4012$}\\
 $5\ (1120 \times 1120)$ & $14.9726$ & $0.4753$ & $0.4235$ & $22.9861$ & $0.6074$ & $0.4018$\\
 $6\ (1344 \times 1344)$ & $14.3093$ & $0.4887$ & $0.4358$ & $23.4824$ & $0.6527$ & $0.383$\\
 $7\ (1568 \times 1568)$ & \makecell{$13.8922$} & \makecell{$0.4852$} & \makecell{$0.4587$} & \makecell{$23.5778$} & \makecell{$0.6588$} & \makecell{$0.3898$}\\
 $8\ (1792 \times 1792)$ & $13.4781$ & $0.4967$ & $0.4656$ & $23.7604$ & $0.70178$ & $0.3638$\\
 $9\ (2016 \times 2016)$ & $13.2869$ & $0.4882$ & $0.4834$ & $23.7907$ & $0.6924$ & $0.3906$\\
 $10\ (2240 \times 2240)$ & \makecell{$13.1013$} & \makecell{$0.4969$} & \makecell{$0.486$} & \makecell{$23.9566$} & \makecell{$0.7244$} & \makecell{$0.3892$}\\
\specialrule{.15em}{.05em}{.05em} 
\end{tabular}}
\end{center}

\vspace{-0.3 cm}
\caption{\label{tab:supp_hires01} Reconstructing upscaled images ($L\in\{1,\dots,10\}$). Upscaled $224 \times 224$ ImageNet samples are reconstructed from standard and AR AlexNet features, the latter predominantly obtaining higher accuracy.}
\vspace{-0.6 cm}
\end{table*}

\subsection{ResNet-18: Robustness Level vs. Reconstruction Accuracy}
\label{sec:supp_inverting_alternative}

We take the ResNet-18 model trained on CIFAR-10 from the \textit{Robustness} library \cite{robustness}, invert its third residual block ($4 \times 4 \times 512$) based on our approach using pixel and feature losses, and evaluate its reconstruction accuracy for standard and AR cases.

We measure the reconstruction accuracy for different robustness levels by training six AR classifiers via $\ell_{2}$ PGD attacks (Madry et al.) with attack radii $\varepsilon$ covering from $0$ to $3.5$ (see \tabref{tab:accuracy02_1}). Accuracy for each model is measured in terms of PSNR, SSIM and LPIPS. We also report the robustness obtained by each model against $\ell_{2}$ PGD attacks.
\begin{table*}[b]
\small
\begin{center}
\resizebox{0.9\textwidth}{!}{
\begin{tabular}{c|c|c|c|c|c|c|c|c} 
\specialrule{.15em}{.05em}{.05em}
&\multicolumn{8}{c}{$\ell_{2}$ PGD Attack ($\varepsilon$)}\\
\cline{2-9}
& $0$ & $0.5$ & $1$ & $1.5$ & $2$ & $2.5$ & $3$ & $3.5$ \\
\hline
\makecell{Standard Accuracy} & $94.93$ & $88.28$ & $81.07$ & $72.47$ & $64.48$ & $64.17$ & $56.77$ & $53.8$\\
\makecell{$\ell_{2}$ PGD Attack} & \makecell{$28.29$\\$(\varepsilon=0.25)$} & \makecell{$68.75$\\$(\varepsilon=0.5)$} & \makecell{$52.24$\\$(\varepsilon=1.0)$} & \makecell{$41.29$\\$(\varepsilon=1.5)$} & \makecell{$34.45$\\$(\varepsilon=2.0)$} & \makecell{$29.63$\\$(\varepsilon=2.5)$} & \makecell{$25.58$\\$(\varepsilon=3.0)$} & \makecell{$23.48$\\$(\varepsilon=3.5)$}\\
\hline
\makecell{PSNR  (dB) $\uparrow$} & $14.7259$ & $18.5161$ & $19.2427$ & $\mathbf{19.6278}$ & ${19.5234}$ & $18.7568$ & $19.3713$  & $19.4376$\\
SSIM $\uparrow$ & $0.2958$ & $0.5179$ & $\mathbf{0.5399}$ & ${0.5332}$ & $0.5265$ & $0.4878$ & $0.501$  & $0.4951$\\
LPIPS $\downarrow$ & $0.6305$ & $0.5024$ & $\mathbf{0.4832}$ & ${0.4905}$ & $0.5019$ & $0.5312$ & $0.5172$ & $0.5321$\\
\specialrule{.15em}{.05em}{.05em}
\end{tabular}}
\end{center}

\vspace{-0.3 cm}
\caption{\label{tab:accuracy02_1} Reconstruction vs. Robustness. ResNet-18 experiments on CIFAR-10 show that learning to invert contracted features with different AR levels significantly affects the reconstruction accuracy.
}
\vspace{-0.6cm}
\end{table*}

Results show the best accuracy is reached for $\varepsilon=1.5$ in terms of PSNR and for $\varepsilon=1$ in terms of SSIM and LPIPS. Quality increases almost monotonically for models with low robustness and reaches a peak of approximately $19.62$ dB PSNR. Models with higher robustness slowly decrease in accuracy, yet obtaining a significant boost over the standard model ($\varepsilon=0$).

\subsection{Comparison Against Alternative Methods}
\label{sec:supp_results_comparison}

\begin{figure*}[t]
\subfloat[Ground-truth images.]{\includegraphics[width=\textwidth]{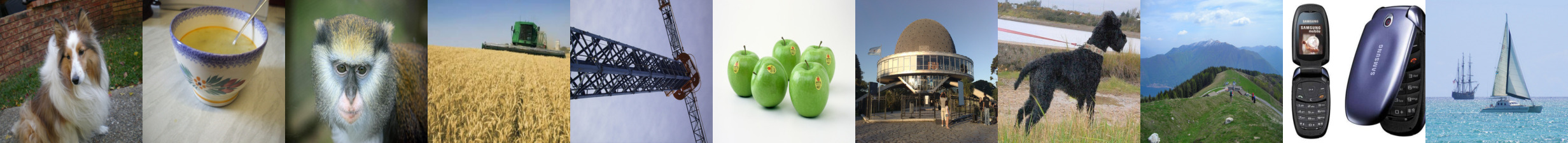}}

\vspace{-.9\baselineskip}
\subfloat[DeePSiM inversion results \cite{dosovitskiy_2016_generating}.]{\includegraphics[width=\textwidth]{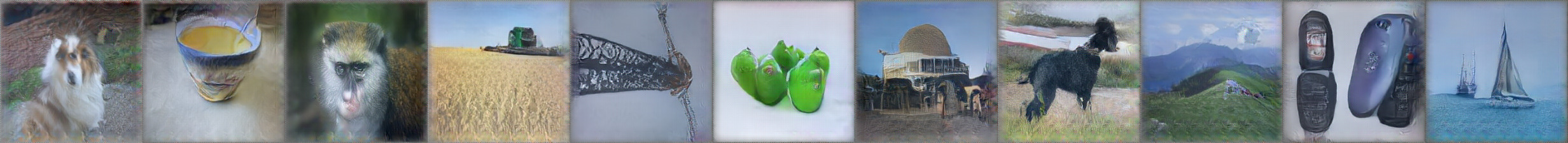}}

\vspace{-.9\baselineskip}
\subfloat[RI results \cite{engstrom_2019_adversarial}.]{\includegraphics[width=\textwidth]{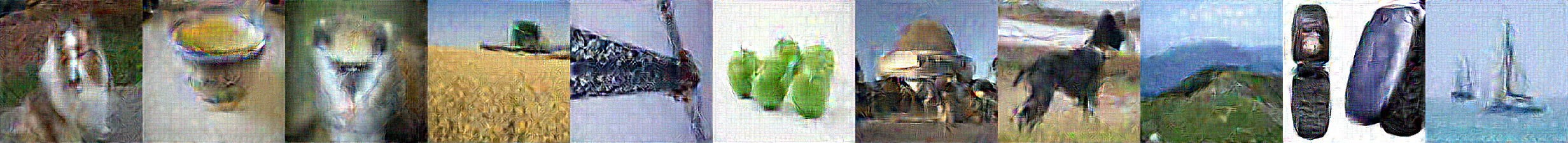}}

\vspace{-.9\baselineskip}
\subfloat[Standard autoencoder.]{\includegraphics[width=\textwidth]{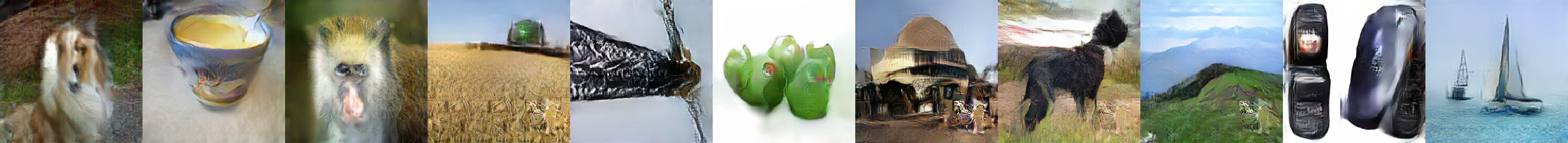}}

\vspace{-.9\baselineskip}
\subfloat[Robust autoencoder (ours).]{\includegraphics[width=\textwidth]{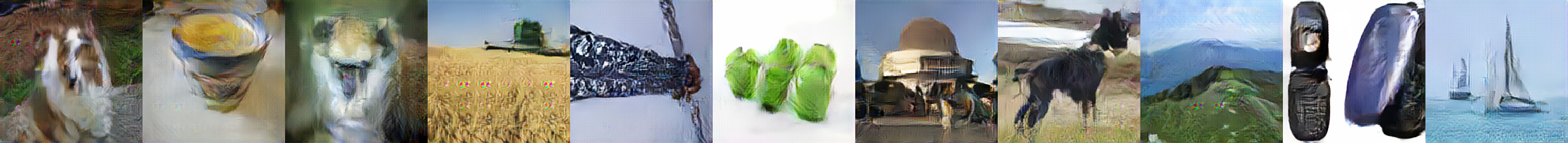}}

\caption{Feature inversion accuracy contrast between our proposed model and alternative inversion methods.}
\label{fig:supp_contrast}
\end{figure*}

Feature inversion accuracy obtained by our proposed model is compared against DeePSiM~\cite{dosovitskiy_2016_generating} and RI~\cite{engstrom_2019_adversarial} methods. \figref{fig:supp_contrast} illustrates the reconstruction accuracy obtained by each method. As previously explained, our generator yields photorealistic results with $37\%$ the trainable parameters required by the DeePSiM generator. Qualitatively, the color distribution obtained by our AR autoencoder is closer to that obtained by DeepSiM. Specifically, without any postprocessing, DeePSiM's results show severe edge distortions, while out method shows minor edge distortions. On the other hand, the optimization based approach from RI introduces several artifacts, despite its use of robust representations. In contrast, our method takes advantage of AR features and minimizes the distortions in a much more efficient manner by replacing the iterative process by a feature inverter (image generator).

Architecture details and training parameters used to train out proposed model are included in \secref{sec:supp_proposed}. DeePSiM results were obtained using its official Caffe implementation. RI results were obtained using its official PyTorch implementation, modified to invert AlexNet \layer{conv5} layer.

\section{Additional Results on Downstream Tasks}
\label{sec:supp_additional}
    \subsection{Style Transfer}
    \label{sec:supp_additional_st}
\begin{figure}[t]
\begin{minipage}[t]{0.1\textwidth}
\centering\textbf{\scriptsize{Refs}}
\end{minipage}\begin{minipage}[t]{0.2\textwidth}
\centering \textbf{\scriptsize{Standard}}
\end{minipage}\begin{minipage}[t]{0.2\textwidth}
\centering \textbf{\scriptsize{AR (ours)}}
\end{minipage}\begin{minipage}[t]{0.1\textwidth}
\centering\textbf{\scriptsize{Refs}}
\end{minipage}\begin{minipage}[t]{0.2\textwidth}
\centering \textbf{\scriptsize{Standard}}
\end{minipage}\begin{minipage}[t]{0.2\textwidth}
\centering \textbf{\scriptsize{AR (ours)}}
\end{minipage}

\vspace{-0.9\baselineskip}
\subfloat[]{\includegraphics[width=0.5\textwidth]{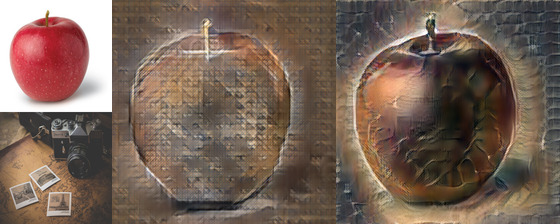}}
\subfloat[]{\includegraphics[width=0.5\textwidth]{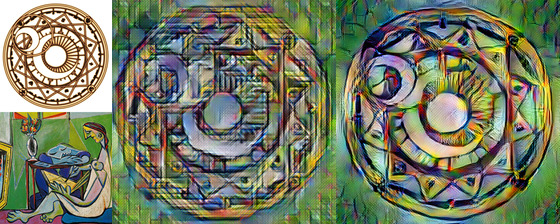}}

\vspace{-.9\baselineskip}
\subfloat[]{\includegraphics[width=0.5\textwidth]{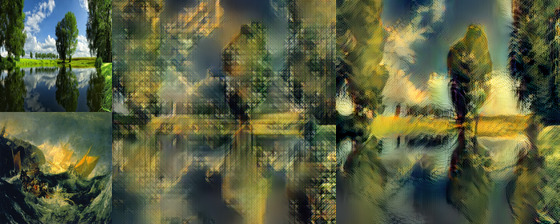}}
\subfloat[]{\includegraphics[width=0.5\textwidth]{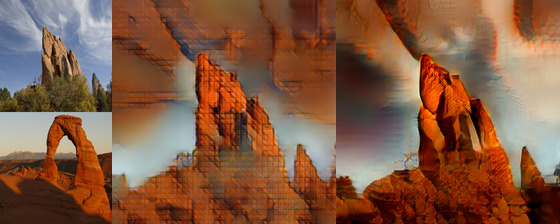}}

\vspace{-.9\baselineskip}
\subfloat[]{\includegraphics[width=0.5\textwidth]{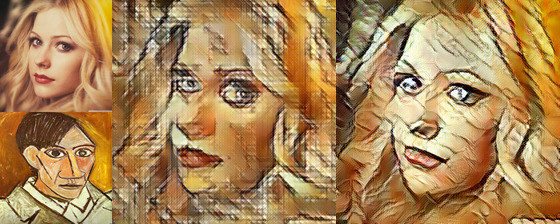}}
\subfloat[]{\includegraphics[width=0.5\textwidth]{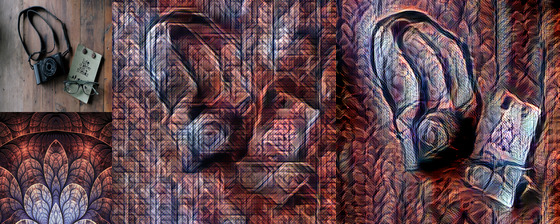}}

\vspace{-.9\baselineskip}
\subfloat[]{\includegraphics[width=0.5\textwidth]{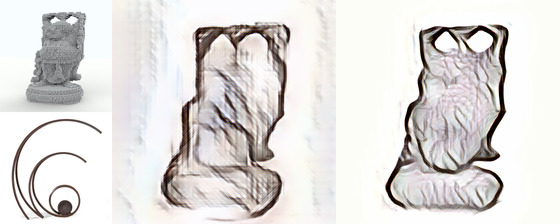}}
\subfloat[]{\includegraphics[width=0.5\textwidth]{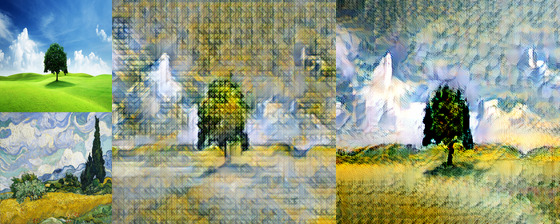}}
\vspace{-0.5\baselineskip}

\caption{Style transfer results using standard and robust AlexNet representations. Stylization obtained using the universal style transfer algorithm \cite{li_2017_universal}.}
\label{fig:supp_st}
\vspace{-0.8 cm}
\end{figure}

    \figref{fig:supp_st} shows additional stylization results obtained via the Universal Style Transfer algorithm using standard and AR AlexNet autoencoders. Qualitatively, the multi-level stylization approach used in our experiments show that AR representations allow a good texture transferring while better preserving the content image structure. Regardless the type of scene being stylized (\eg landscapes, portraits or single objects), aligning AR robust features allows to preserve sharp edges and alleviates the distortions generated by aligning standard features. Architecture details and training parameters for the style transfer experiments are covered in \secref{sec:supp_proposed_st}.

    \begin{figure*}[t]

\begin{minipage}{0.25\textwidth}
\centering\textbf{\colorbox{white}{\scalebox{.7}{Ground-truth}}}
\end{minipage}\begin{minipage}{0.25\textwidth}
\centering \textbf{\colorbox{white}{\scalebox{.7}{Noisy}}}
\end{minipage}\begin{minipage}{0.25\textwidth}
\centering \textbf{\colorbox{white}{\scalebox{.7}{Standard}}}
\end{minipage}\begin{minipage}{0.25\textwidth}
\centering\textbf{\colorbox{white}{\scalebox{.7}{Robust}}}
\end{minipage}

\vspace{-\baselineskip}
\subfloat{\includegraphics[width=\textwidth]{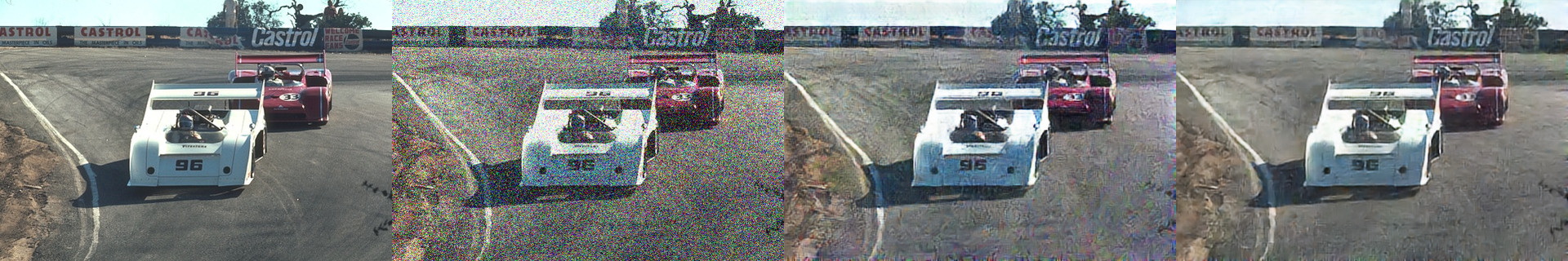}}

\vspace{-0.8\baselineskip}
\subfloat{\includegraphics[width=\textwidth]{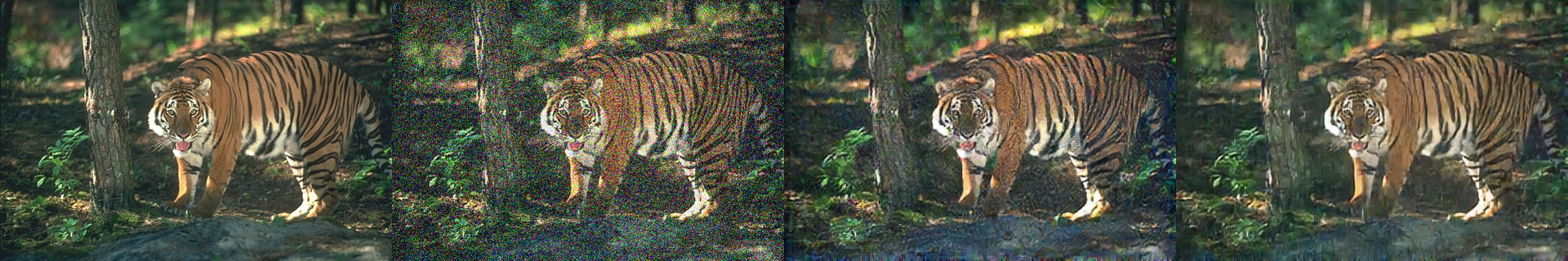}}

\vspace{-0.8\baselineskip}
\subfloat{\includegraphics[width=\textwidth]{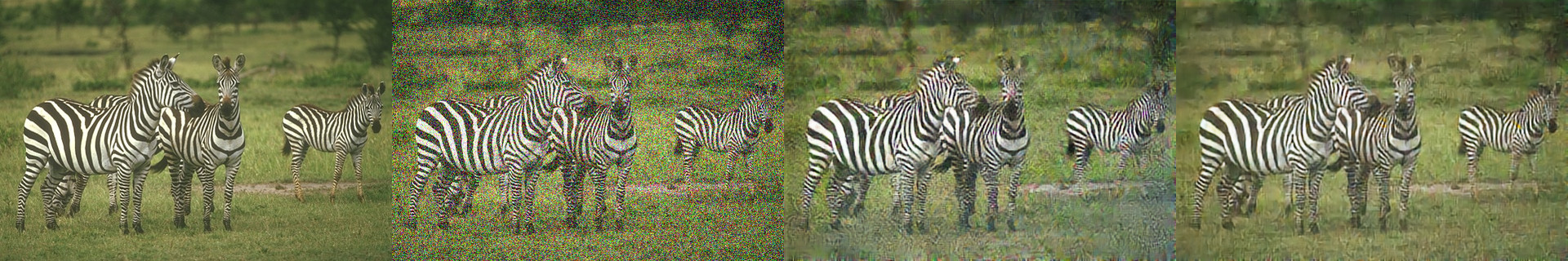}}

\vspace{-0.8\baselineskip}
\subfloat{\includegraphics[width=\textwidth]{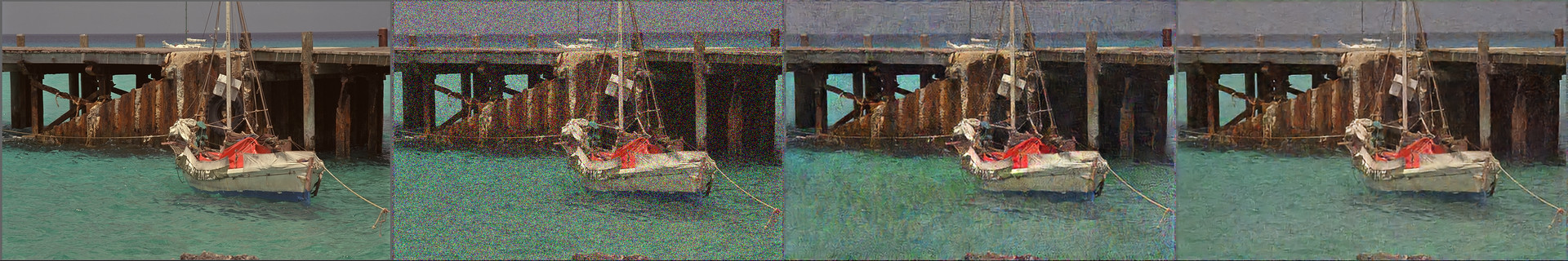}}

\vspace{-0.8\baselineskip}
\subfloat{\includegraphics[width=\textwidth]{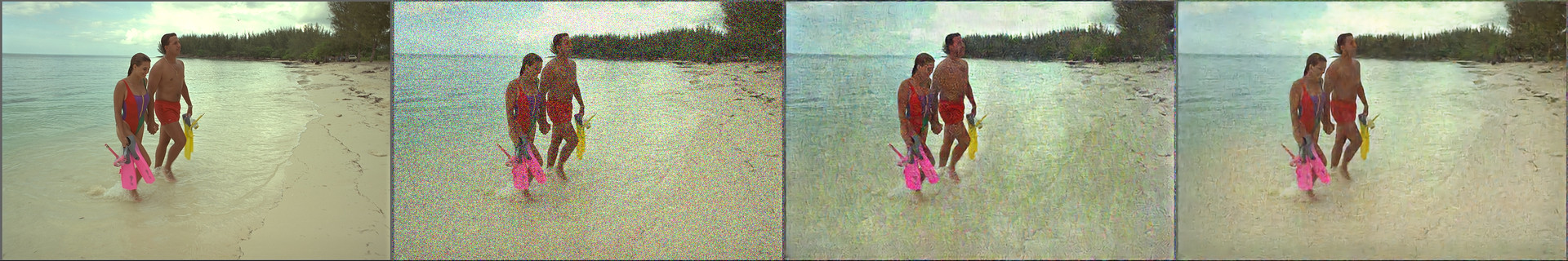}}

\vspace{-0.8\baselineskip}
\subfloat{\includegraphics[width=\textwidth]{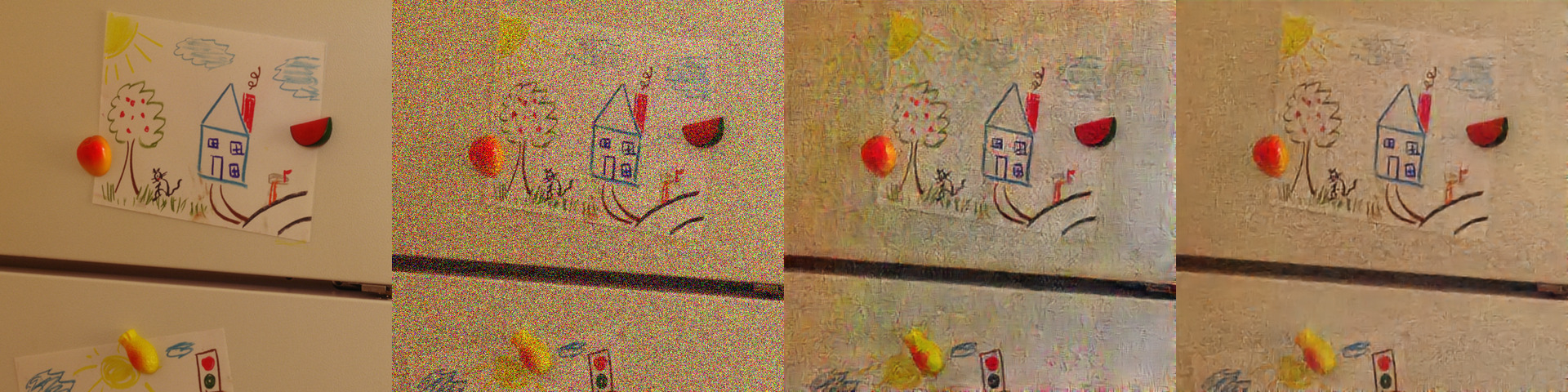}}

\vspace{-0.8\baselineskip}
\subfloat{\includegraphics[width=\textwidth]{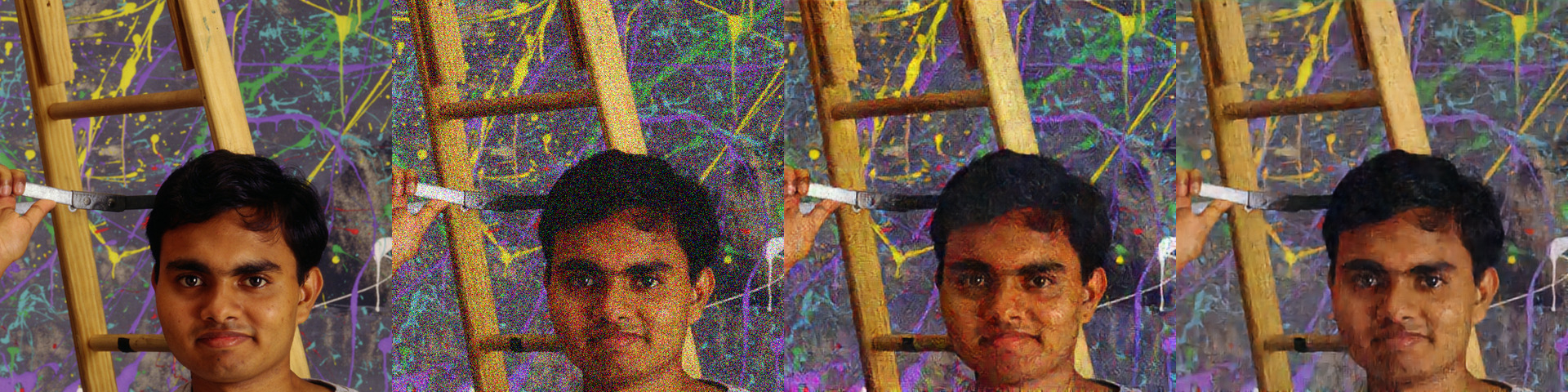}}

\vspace{-0.3cm}
\caption{Image denoising results using standard and AR encoders (AlexNet) from the CBSD68 and Kodak24 sets. Samples corrupted by clipped white Gaussian noise $(\sigma= \frac{50}{255})$.}
\label{fig:supp_denoising}
\vspace{-0.6cm}
\end{figure*}

    \subsection{Image Denoising}
    \label{sec:supp_additional_denoising}
    \figref{fig:supp_denoising} shows additional denoising results using our standard and AR autoencoders for the CBSDS68, Kodak24 and McMaster datasets. As previously discussed, we leverage the low-level feature representations by adding skip connections to our proposed autoencoder. Low-level features complement the contracted feature map obtained from AlexNet \layer{conv5}, improving the detail preservation. This is observed in the results, both with standard and AR autoencoders.
    
    On the other hand, despite the effect of using skip connections, reconstructions from AR representations show a notorious improvement with respect to standard reconstructions. Specifically, by combining skip connections with the rich information already encapsulated in robust representations, results on all three datasets show a substantial denoising improvement.

\section{Implementation Details}\label{sec:supp_proposed_method}
    \subsection{Architecture and Training Details}
    \label{sec:supp_proposed}
    
    \textbf{Encoder.} For all downstream tasks, our adversarially robust AlexNet classifier was trained using PGD attacks \cite{madry_2018_towards}. The process was performed on ImageNet using stochastic gradient descent. The AR training parameters are as follows:
    \begin{itemize}
        \item Perturbation constraint: $\ell_{2}$ ball with $\varepsilon=3$
        \item PGD attack steps: $7$
        \item Step size: $0.5$
        \item Training epochs: $90$
    \end{itemize}
    
    On the other hand, the standard AlexNet classifier was trained using cross-entropy loss as optimization criteria. For both cases, the training parameters were the following:
    \begin{itemize}
        \item Initial learning rate: $0.1$
        \item Optimizer: Learn rate divided by a factor of $10$ every $30$ epochs.
        \item Batch size: $256$
    \end{itemize}

    Tested under AutoAttack ($\ell_{2}, \varepsilon=3$), our AR AlexNet obtains a $18.7\%$ top-1 robust accuracy, while our standard AlexNet classifier obtains a $0\%$ top-1 robust accuracy.
    
    AR training was performed using the \textit{Robustness} library \cite{robustness} on four Tesla V100 GPUs. Additional details about the model architecture and training parameters used for each experiment and downstream task are as follows.
    
    \textbf{Feature Inversion Experiments.} A fully convolutional architecture is used for the decoder or image generator. \tabref{tab:supp_generator01} describes the decoder architecture used to invert both standard and AR representations, where \texttt{conv2d} denotes a $2$D convolutional layer, \texttt{tconv2d} a $2$D transposed convolutional layer, \texttt{BN} batch normalization, \texttt{ReLU} the rectified linear unit operator and \texttt{tanh} the hyperbolic tangent operator.
    
    \begin{figure}[t]
\vspace{-1 cm}
\centering
\subfloat[]{\includegraphics[width=0.9\textwidth]{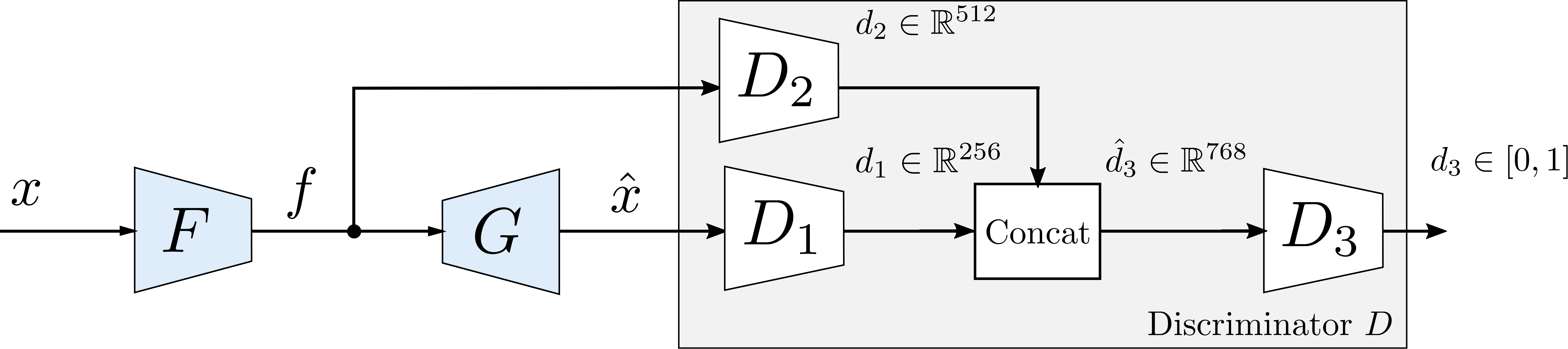}}

\vspace{-0.3cm}
\caption{Discriminator model.}
\label{fig:supp_disc_diag}
\vspace{-0.6cm}
\end{figure}

    \tabref{tab:supp_discriminator01} shows the discriminator architecture, where \texttt{leakyReLU} corresponds to the leaky rectified linear unit, \texttt{linear} to a fully-connected layer, \texttt{apooling} to average pooling and \texttt{sigmoid} to the Sigmoid operator. Motivated by the architecture proposed by Dosovitskiy \& Brox \cite{dosovitskiy_2016_generating}, the discriminator takes as input both a real or fake image and its target \layer{conv5} feature map to compute the probability of the sample being real. \figref{fig:supp_disc_diag} shows the discriminator architecture.

    Standard and AR autoencoders were trained on ImageNet using $\ell_{1}$ pixel, feature and GAN losses using ADAM. In both cases, all convolutional and transposed convolutional layers are regularized using spectral normalization \cite{miyato_2018_spectral}. Training was performed using Pytorch-only code on two Tesla V100 GPUs.
    
    The loss weights and training setup for both standard and AR cases correspond to:
    \begin{itemize}
     \setlength\itemsep{0.1\baselineskip}
        \item Generator weights: $\lambda_{\text{pix}}=2\times 10^{-6}, \lambda_{\text{feat}}=1\times 10^{-2}, \lambda_{\text{GAN}}= 100$
        \item Discriminator weight: $\lambda_{\text{disc}}=2\times 10^{-6}$
        \item Training epochs: $90$
        \item Generator initial learning rate: $3\times 10^{-4}$ (divided by a factor of $10$ every $30$ epochs).
        \item Discriminator initial learning rate: $12\times 10^{-4}$ (divided by a factor of $10$ every $30$ epochs).
        \item LeakyReLU factor: $0.2$
        \item ADAM $\beta\in [0, 0.9]$
        \item Batch size: $128$
    \end{itemize}
    
    \begin{table*}[t]
\begin{center}
\resizebox{0.9\textwidth}{!}{
\begin{tabular}{c|c|c|c|c|c|c|c|c|c}
\specialrule{.15em}{.05em}{.05em} 
\makecell{Layer}
& \makecell{Layer Type}
& \makecell{Kernel\\ Size}
& \makecell{Bias}
& \makecell{Stride}
& \makecell{Pad}
& \makecell{Input\\ Size}
& \makecell{Output\\ Size}
& \makecell{Input\\ Channels}
& \makecell{Output\\ Channels} \\
\midrule
1a & conv2d + BN + ReLU & $3\times 3$ & \xmark & $1$ & $1$ & $6\times 6$ & $6\times 6$ & $256$ & $256$ \\
\midrule
2a & tconv2d + BN + ReLU & $4\times 4$ & \xmark & $1$ & $1$ & $6\times 6$ & $7\times 7$ & $256$ & $256$ \\
2b & conv2d + BN + ReLU & $3\times 3$ & \xmark & $1$ & $1$ & $7\times 7$ & $7\times 7$ & $256$ & $256$ \\
\midrule
3a & tconv2d + BN + ReLU & $4\times 4$ & \xmark & $2$ & $1$ & $7\times 7$ & $14\times 14$ & $256$ & $256$ \\
3b & conv2d + BN + ReLU & $3\times 3$ & \xmark & $1$ & $1$ & $14\times 14$ & $14\times 14$ & $256$ & $256$ \\
\midrule
4a & tconv2d + BN + ReLU & $4\times 4$ & \xmark & $2$ & $1$ & $14\times 14$ & $28\times 28$ & $256$ & $256$ \\
4b & conv2d + BN + ReLU & $3\times 3$ & \xmark & $1$ & $1$ & $28\times 28$ & $28\times 28$ & $256$ & $128$ \\
\midrule
5a & tconv2d + BN + ReLU & $4\times 4$ & \xmark & $2$ & $1$ & $28\times 28$ & $56\times 56$ & $128$ & $128$ \\
5b & conv2d + BN + ReLU & $3\times 3$ & \xmark & $1$ & $1$ & $56\times 56$ & $56\times 56$ & $128$ & $64$ \\
\midrule
6a & tconv2d + BN + ReLU & $4\times 4$ & \xmark & $2$ & $1$ & $56\times 56$ & $112\times 112$ & $64$ & $64$ \\
6b & conv2d + BN + ReLU & $3\times 3$ & \xmark & $1$ & $1$ & $112\times 112$ & $112\times 112$ & $64$ & $32$ \\
\midrule
7a & tconv2d + BN + ReLU & $4\times 4$ & \xmark & $2$ & $1$ & $112\times 112$ & $224\times 224$ & $32$ & $32$ \\
7b & conv2d + BN + ReLU & $3\times 3$ & \xmark & $1$ & $1$ & $224\times 224$ & $224\times 224$ & $32$ & $3$ \\
7c & conv2d + tanh & $3\times 3$ & \cmark & $1$ & $1$ & $224\times 224$ & $224\times 224$ & $3$ & $3$\\
\specialrule{.15em}{.05em}{.05em}
\end{tabular}}
\end{center}
\vspace{-0.3 cm}
\caption{\label{tab:supp_generator01} Generator architecture used for feature inversion.}
\vspace{-0.6 cm}
\end{table*}

    \begin{table*}[t]
\begin{center}
\resizebox{0.9\textwidth}{!}{
\begin{tabular}{c|c|c|c|c|c|c|c|c|c}
\specialrule{.15em}{.05em}{.05em} 
\makecell{Layer}
& \makecell{Layer Type}
& \makecell{Kernel\\ Size}
& \makecell{Bias}
& \makecell{Stride}
& \makecell{Pad}
& \makecell{Input\\ Size}
& \makecell{Output\\ Size}
& \makecell{Input\\ Channels}
& \makecell{Output\\ Channels} \\
\specialrule{.15em}{.05em}{.05em}
\multicolumn{10}{c}{Feature Extractor 1 ($D_{1}$)}\\
\midrule
1a & conv2d + ReLU & $3\times 3$ & \cmark & $4$ & $1$ & $256\times 256$ & $56\times 56$ & $3$ & $32$ \\
\midrule
2a & conv2d + ReLU & $5\times 5$ & \cmark & $1$ & $1$ & $56\times 56$ & $52\times 52$ & $32$ & $64$ \\
2b & conv2d + ReLU & $3\times 3$ & \cmark & $2$ & $1$ & $52\times 52$ & $23\times 23$ & $64$ & $128$ \\
\midrule
3a & conv2d + ReLU & $3\times 3$ & \cmark & $1$ & $1$ & $23\times 23$ & $21\times 21$ & $128$ & $256$ \\
3b & conv2d + ReLU & $3\times 3$ & \cmark & $2$ & $1$ & $21\times 21$ & $11\times 11$ & $256$ & $256$ \\
\midrule
4 & ave. pooling & $11 \times 11$ & $-$ & $-$ & $-$ & $11\times 11$ & $1 \times 1$ & $256$ & $256$\\
\specialrule{.15em}{.05em}{.05em}
\multicolumn{10}{c}{Classifier 1 ($D_{2}$)}\\
\midrule
4a & Linear + ReLU & $-$ & \cmark & $-$ & $1$ & $9216$ & $1024$ & $-$ & $-$ \\
4b & Linear + ReLU & $-$ & \cmark & $-$ & $1$ & $1024$ & $512$ & $-$ & $-$ \\
\specialrule{.15em}{.05em}{.05em}
\multicolumn{10}{c}{Classifier 2 ($D_{3}$)}\\
\midrule
5a & Linear + ReLU & $-$ & \cmark & $-$ & $1$ & $768$ & $512$ & $-$ & $-$ \\
5b & Linear + Sigmoid & $-$ & \cmark & $-$ & $1$ & $512$ & $1$ & $-$ & $-$ \\
\specialrule{.15em}{.05em}{.05em}
\end{tabular}}
\end{center}
\vspace{-0.3 cm}
\caption{\label{tab:supp_discriminator01} Discriminator architecture used for feature inversion.}
\vspace{-0.6 cm}
\end{table*}

    \subsection{Style Transfer}
    \label{sec:supp_proposed_st}
    
    While, for standard and AR scenarios, the autoencoder associated to \layer{conv5} corresponds to the model described in \secref{sec:supp_proposed}, those associated to \layer{conv1} and \layer{conv2} use Nearest neighbor interpolation instead of transposed convolution layers to improve the reconstruction accuracy and to avoid the checkerboard effect generated by transposed convolutional layers. \tabref{tab:supp_st01}, and \tabref{tab:supp_st02} describe their architecture details.
    
    All generators were fully-trained on ImageNet using Pytorch-only code on two Tesla V100 GPUs. The regularization parameters and training setup for both cases are as follows:
    \begin{itemize}
     \setlength\itemsep{0.1\baselineskip}
        \item Standard generator weights: $\lambda_{\text{pix}}=2\times 10^{-4}, \lambda_{\text{feat}}=1\times 10^{-2}$.
        \item AR generator weights: $\lambda_{\text{pix}}=2\times 10^{-6}, \lambda_{\text{feat}}=1\times 10^{-2}$.
        \item Training epochs: $90$.
        \item Generator initial learning rate: $3\times 10^{-4}$ (divided by a factor of $10$ every $30$ epochs).
        \item ADAM $\beta\in [0, 0.9]$.
        \item Batch size: $128$.
    \end{itemize}
    
    \begin{table*}[t]
\begin{center}
\resizebox{0.9\textwidth}{!}{
\begin{tabular}{c|c|c|c|c|c|c|c|c|c}
\specialrule{.15em}{.05em}{.05em} 
\makecell{Layer}
& \makecell{Layer Type}
& \makecell{Kernel\\ Size}
& \makecell{Bias}
& \makecell{Stride}
& \makecell{Pad}
& \makecell{Input\\ Size}
& \makecell{Output\\ Size}
& \makecell{Input\\ Channels}
& \makecell{Output\\ Channels} \\
\midrule
1a & conv2d + BN + ReLU & $3\times 3$ & \xmark & $1$ & $1$ & $27\times 27$ & $27\times 27$ & $64$ & $64$ \\
\midrule
2a & tconv2d + BN + ReLU & $4\times 4$ & \xmark & $1$ & $1$ & $27\times 27$ & $28\times 28$ & $64$ & $64$ \\
2b & conv2d + BN + ReLU & $3\times 3$ & \xmark & $1$ & $1$ & $28\times 28$ & $28\times 28$ & $64$ & $64$ \\
\midrule
3a & NN interpolation & $-$ & $-$ & $2$ & $-$ & $28\times 28$ & $56\times 56$ & $64$ & $64$ \\
3b & conv2d + BN + ReLU & $3\times 3$ & \xmark & $1$ & $1$ & $56\times 56$ & $56\times 56$ & $64$ & $64$ \\
3c & conv2d + BN + ReLU & $3\times 3$ & \xmark & $1$ & $1$ & $56\times 56$ & $56\times 56$ & $64$ & $32$ \\
\midrule
4a & NN interpolation & $-$ & $-$ & $2$ & $-$ & $56\times 56$ & $112\times 112$ & $32$ & $32$ \\
4b & conv2d + BN + ReLU & $3\times 3$ & \xmark & $1$ & $1$ & $112 \times 112$ & $112\times 112$ & $32$ & $32$ \\
\midrule
5a & NN interpolation & $-$ & $-$ & $2$ & $-$ & $112\times 112$ & $224\times 224$ & $32$ & $32$ \\
5b & conv2d + BN + ReLU & $3\times 3$ & \xmark & $1$ & $1$ & $224\times 224$ & $224\times 224$ & $32$ & $16$ \\
5c & conv2d + BN + ReLU & $3\times 3$ & \xmark & $1$ & $1$ & $224\times 224$ & $224\times 224$ & $16$ & $3$ \\
5d & conv2d + tanh & $3\times 3$ & \cmark & $1$ & $1$ & $224\times 224$ & $224\times 224$ & $3$ & $3$\\
\specialrule{.15em}{.05em}{.05em}
\end{tabular}}
\end{center}
\vspace{-0.3 cm}
\caption{\label{tab:supp_st01} \layer{Conv1} generator architecture used for style transfer.}
\vspace{-0.6 cm}
\end{table*}

    \begin{table*}[t]
\begin{center}
\resizebox{0.9\textwidth}{!}{
\begin{tabular}{c|c|c|c|c|c|c|c|c|c}
\specialrule{.15em}{.05em}{.05em} 
\makecell{Layer}
& \makecell{Layer Type}
& \makecell{Kernel\\ Size}
& \makecell{Bias}
& \makecell{Stride}
& \makecell{Pad}
& \makecell{Input\\ Size}
& \makecell{Output\\ Size}
& \makecell{Input\\ Channels}
& \makecell{Output\\ Channels} \\
\midrule
1a & conv2d + BN + ReLU & $3\times 3$ & \xmark & $1$ & $1$ & $13\times 13$ & $13\times 13$ & $192$ & $192$ \\
\midrule
2a & tconv2d + BN + ReLU & $4\times 4$ & \xmark & $1$ & $1$ & $13\times 13$ & $14\times 14$ & $192$ & $192$ \\
2b & conv2d + BN + ReLU & $3\times 3$ & \xmark & $1$ & $1$ & $14\times 14$ & $14\times 14$ & $192$ & $96$ \\
\midrule
3a & NN interpolation & $-$ & $-$ & $2$ & $-$ & $14\times 14$ & $28\times 28$ & $96$ & $96$ \\
3b & conv2d + BN + ReLU & $3\times 3$ & \xmark & $1$ & $1$ & $28\times 28$ & $28\times 28$ & $96$ & $96$ \\
3c & conv2d + BN + ReLU & $3\times 3$ & \xmark & $1$ & $1$ & $28\times 28$ & $28\times 28$ & $96$ & $64$ \\
\midrule
4a & NN interpolation & $-$ & $-$ & $2$ & $-$ & $28\times 28$ & $56\times 56$ & $64$ & $64$ \\
4b & conv2d + BN + ReLU & $3\times 3$ & \xmark & $1$ & $1$ & $56 \times 56$ & $56\times 56$ & $64$ & $64$ \\
\midrule
5a & NN interpolation & $-$ & $-$ & $2$ & $-$ & $56\times 56$ & $112\times 112$ & $64$ & $64$ \\
5b & conv2d + BN + ReLU & $3\times 3$ & \xmark & $1$ & $1$ & $112\times 112$ & $112\times 112$ & $64$ & $64$ \\
\midrule
6a & NN interpolation & $-$ & $-$ & $2$ & $-$ & $112\times 112$ & $224\times 224$ & $64$ & $64$ \\
6b & conv2d + BN + ReLU & $3\times 3$ & \xmark & $1$ & $1$ & $224\times 224$ & $224\times 224$ & $64$ & $32$ \\
6c & conv2d + BN + ReLU & $3\times 3$ & \xmark & $1$ & $1$ & $224\times 224$ & $224\times 224$ & $32$ & $3$ \\
6d & conv2d + tanh & $3\times 3$ & \cmark & $1$ & $1$ & $224\times 224$ & $224\times 224$ & $3$ & $3$\\
\specialrule{.15em}{.05em}{.05em}
\end{tabular}}
\end{center}
\vspace{-0.3 cm}
\caption{\label{tab:supp_st02} \layer{Conv2} generator architecture used for style transfer.}
\vspace{-0.6 cm}
\end{table*}

    \subsection{Image Denoising}
    \label{sec:supp_proposed_denoising}
   
\begin{figure}[t]
\vspace{-1 cm}
\centering
\subfloat[Skip connected AlexNet encoder]{\includegraphics[height=0.09\textheight]{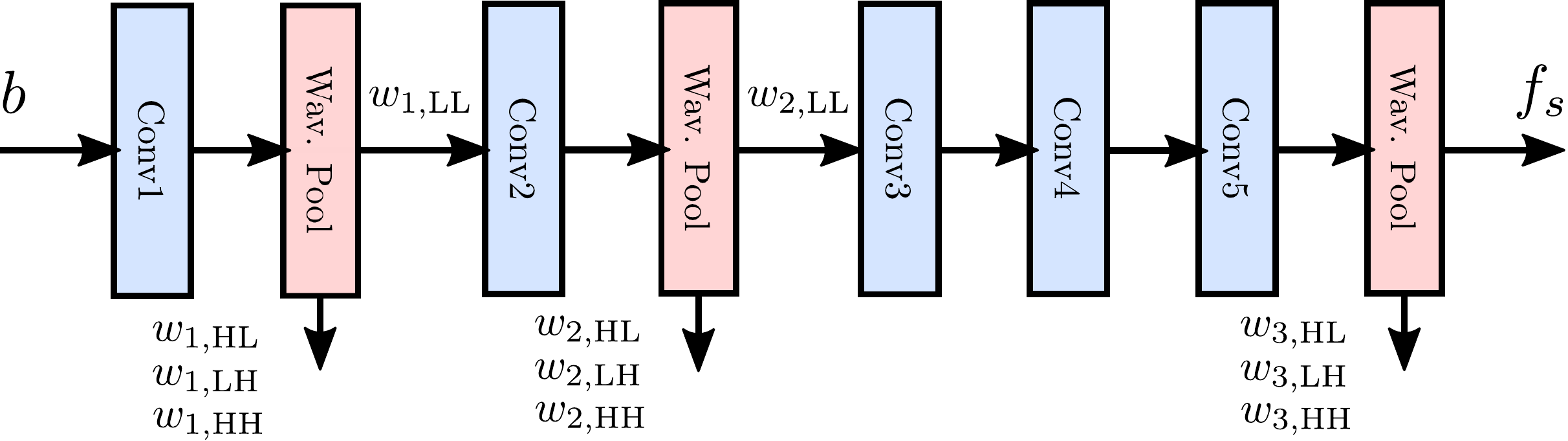}}

\subfloat[Skip connected AlexNet decoder]{\includegraphics[width=1\textwidth]{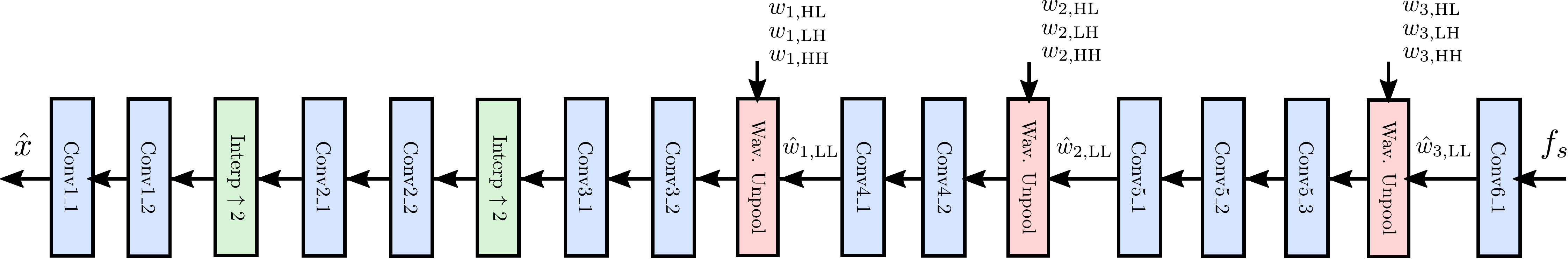}}

\vspace{-0.3cm}
\caption{Proposed denoising autoencoder including skip connections.}
\label{fig:supp_den_diag}
\vspace{-0.6cm}
\end{figure}

    Our image denoising model consists of standard and AR autoencoders equipped with skip connections to better preserve image details. \figref{fig:supp_den_diag} illustrates the proposed denoising model, where skip connections follow the Wavelet Pooling approach \cite{yoo_2019_photorealistic}. \tabref{tab:supp_den_enc} and \tabref{tab:supp_den_dec} include additional encoder and decoder architecture details, respectively.

    Encoder pooling layers are replaced by Haar wavelet analysis operators, generating an approximation component, denoted as $\{w_{k, \text{LL}}\}$, and three detail components, denoted as $\{w_{k, \text{LH}}, w_{k, \text{HL}}, w_{k, \text{HH}}\}$, where $k$ corresponds to the pooling level. While the approximation (low-frequency) component is passed to the next encoding layer, details are skip-connected to their corresponding stages in the decoder. Following this, transposed convolutional layers in the decoder are replaced by unpooling layers (Haar wavelet synthesis operators), reconstructing a signal with well-preserved details at each level and improving reconstruction.
    
    In contrast to the AlexNet architecture, all convolutional layers on the decoder use kernels of size $3 \times 3$. Also, given the striding factor of the first two AlexNet convolutional layers, two additional interpolation layers of striding factor $2$ are used to recover the original input size ($224 \times 224$).
    
    Standard and AR robust generators were trained using exclusively $\ell_{1}$ pixel and feature losses. Training was performed on ImageNet using Pytorch-only code on four Tesla V100 GPUs. Generator loss weights and training parameters for both cases correspond to:
    \begin{itemize}
     \setlength\itemsep{0.1\baselineskip}
        \item Generator weights: $\lambda_{\text{pix}}=2\times 10^{-6}, \lambda_{\text{feat}}=1\times 10^{-2}$.
        \item Training epochs: $90$.
        \item Generator initial learning rate: $3\times 10^{-4}$ (divided by a factor of $10$ every $30$ epochs).
        \item ADAM $\beta\in [0, 0.9]$.
        \item Batch size: $128$.
    \end{itemize}
    \begin{table*}[t]
\begin{center}
\resizebox{0.9\textwidth}{!}{
\begin{tabular}{c|c|c|c|c|c|c|c|c|c}
\specialrule{.15em}{.05em}{.05em} 
\makecell{Layer}
& \makecell{Layer Type}
& \makecell{Kernel\\ Size}
& \makecell{Bias}
& \makecell{Stride}
& \makecell{Pad}
& \makecell{Input\\ Size}
& \makecell{Output\\ Size}
& \makecell{Input\\ Channels}
& \makecell{Output\\ Channels} \\
\midrule
1a & conv2d + ReLU & $11\times 11$ & \cmark & $4$ & $2$ & $224\times 224$ & $55\times 55$ & $3$ & $64$ \\
\midrule
2a & Wavelet pooling & $-$ & $-$ & $2$ & $-$ & $55\times 55$ & $27\times 27$ & $64$ & $64$ \\
2b & conv2d + ReLU & $5\times 5$ & \cmark & $1$ & $2$ & $27\times 27$ & $27\times 27$ & $64$ & $192$ \\
\midrule
3a & Wavelet pooling & $-$ & $-$ & $2$ & $-$ & $27\times 27$ & $13\times 13$ & $192$ & $192$ \\
3b & conv2d + ReLU & $3\times 3$ & \cmark & $1$ & $1$ & $13\times 13$ & $13\times 13$ & $192$ & $384$ \\
3c & conv2d + ReLU & $3\times 3$ & \cmark & $1$ & $1$ & $13\times 13$ & $13\times 13$ & $384$ & $256$ \\
3c & conv2d + ReLU & $3\times 3$ & \cmark & $1$ & $1$ & $13\times 13$ & $13\times 13$ & $256$ & $256$ \\
\midrule
4a & Wavelet pooling & $-$ & $-$ & $2$ & $-$ & $13\times 13$ & $6\times 6$ & $256$ & $256$ \\
\specialrule{.15em}{.05em}{.05em}
\end{tabular}}
\end{center}
\vspace{-0.3 cm}
\caption{\label{tab:supp_den_enc} Encoder architecture used for image denoising.}
\vspace{-0.6 cm}
\end{table*}

    \begin{table*}[t]
\begin{center}
\resizebox{0.9\textwidth}{!}{
\begin{tabular}{c|c|c|c|c|c|c|c|c|c}
\specialrule{.15em}{.05em}{.05em} 

& \makecell{Layer Type}
& \makecell{Kernel\\ Size}
& \makecell{Bias}
& \makecell{Stride}
& \makecell{Pad}
& \makecell{Input\\ Size}
& \makecell{Output\\ Size}
& \makecell{Input\\ Channels}
& \makecell{Output\\ Channels} \\
\midrule
1a & conv2d + BN + ReLU & $3\times 3$ & \xmark & $1$ & $1$ & $6\times 6$ & $6\times 6$ & $256$ & $256$ \\
\midrule
2a & Wavelet unpooling & $-$ & $-$ & $2$ & $-$ & $6\times 6$ & $12\times 12$ & $256$ & $256$ \\
2b & conv2d + BN + ReLU & $3\times 3$ & \xmark & $1$ & $1$ & $12\times 12$ & $12\times 12$ & $256$ & $256$ \\
2c & Reflection padding & $-$ & $-$ & $-$ & $-$ & $12\times 12$ & $13\times 13$ & $256$ & $256$ \\
2d & conv2d + BN + ReLU & $3\times 3$ & \xmark & $1$ & $1$ & $13\times 13$ & $13\times 13$ & $256$ & $256$ \\
2e & conv2d + BN + ReLU & $3\times 3$ & \xmark & $1$ & $1$ & $13\times 13$ & $13\times 13$ & $256$ & $192$ \\
\midrule
3a & Wavelet unpooling & $-$ & $-$ & $2$ & $-$ & $13\times 13$ & $26\times 26$ & $192$ & $192$ \\
3b & Reflection padding & $-$ & $-$ & $-$ & $-$ & $26\times 26$ & $27\times 27$ & $192$ & $192$ \\
3c & conv2d + BN + ReLU & $3\times 3$ & \xmark & $1$ & $1$ & $27\times 27$ & $27\times 27$ & $192$ & $128$ \\
3d & conv2d + BN + ReLU & $3\times 3$ & \xmark & $1$ & $1$ & $27\times 27$ & $27\times 27$ & $128$ & $64$ \\
\midrule
4a & Wavelet unpooling & $-$ & $-$ & $2$ & $-$ & $27\times 27$ & $55\times 55$ & $64$ & $64$ \\
4b & Reflection padding & $-$ & $-$ & $-$ & $-$ & $55\times 55$ & $56\times 56$ & $64$ & $64$ \\
4c & conv2d + BN + ReLU & $3\times 3$ & \xmark & $1$ & $1$ & $56 \times 56$ & $56\times 56$ & $64$ & $64$ \\
\midrule
5a & NN interpolation & $-$ & $-$ & $2$ & $-$ & $56\times 56$ & $112\times 112$ & $64$ & $64$ \\
5b & conv2d + BN + ReLU & $3\times 3$ & \xmark & $1$ & $1$ & $112 \times 112$ & $112\times 112$ & $64$ & $32$ \\
5c & conv2d + BN + ReLU & $3\times 3$ & \xmark & $1$ & $1$ & $112 \times 112$ & $112\times 112$ & $32$ & $32$ \\
\midrule
6a & NN interpolation & $-$ & $-$ & $2$ & $-$ & $112\times 112$ & $224\times 224$ & $32$ & $32$ \\
6b & conv2d + BN + ReLU & $3\times 3$ & \xmark & $1$ & $1$ & $224\times 224$ & $224\times 224$ & $32$ & $3$ \\
6c & conv2d + BN + ReLU & $3\times 3$ & \xmark & $1$ & $1$ & $224\times 224$ & $224\times 224$ & $3$ & $3$ \\
6d & conv2d + tanh & $3\times 3$ & \cmark & $1$ & $1$ & $224\times 224$ & $224\times 224$ & $3$ & $3$\\
\specialrule{.15em}{.05em}{.05em}
\end{tabular}}
\end{center}
\caption{\label{tab:supp_den_dec} Decoder architecture used for image denoising.}
\end{table*}

\end{document}